\documentclass[10pt,twocolumn,letterpaper]{article}

\usepackage[pagenumbers]{cvpr} 

\usepackage{graphicx}
\usepackage{subcaption}
\usepackage{amsmath}
\usepackage{amssymb}
\usepackage{booktabs}
\usepackage[ruled,linesnumbered]{algorithm2e}
\usepackage{multirow}

\usepackage[table]{xcolor}

\newcommand{\camera}[1]{\textcolor{black}{#1}}
\newcommand{\keypoint}[1]{\vspace{0.05cm}\noindent\textbf{#1}}
\def\eg{\emph{e.g.~}} 
\def\ie{\emph{i.e.~}} 
\newcommand{\sgg}[1]{{\color{black}#1}}
\newcommand{\new}[1]{{\color{black}#1}}

\usepackage[pagebackref,breaklinks,colorlinks]{hyperref}
\usepackage{appendix}

\usepackage[accsupp]{axessibility}  

\usepackage[capitalize]{cleveref}
\crefname{section}{Sec.}{Secs.}
\Crefname{section}{Section}{Sections}
\Crefname{table}{Table}{Tables}
\crefname{table}{Tab.}{Tabs.}

\begin{document}

\title{Ranking Distance Calibration for Cross-Domain Few-Shot Learning}



\author{Pan Li$^{1}$, Shaogang Gong$^{1}$, Chengjie Wang$^{2}$ and Yanwei Fu$^{3}$ \\
$^1$Queen Mary University of London \quad
$^2$Tencent Youtu Lab\quad
$^3$Fudan University \\
{\tt\small \{pan.li, s.gong\}@qmul.ac.uk,\quad}
{\tt\small jasoncjwang@tencent.com,\quad
yanweifu@fudan.edu.cn} 
}

\maketitle

\begin{abstract}
Recent progress in few-shot learning promotes a more realistic cross-domain setting, where the source and target datasets are in different domains. 
Due to the domain gap and disjoint label \sgg{spaces} between source and target datasets, their shared knowledge is extremely limited.
This encourages us to \sgg{explore more information in the target domain rather than to overly} elaborate training \sgg{strategies} on the source \sgg{domain as} in many existing methods.
\sgg{Hence}, we start from a \sgg{generic representation pre-trained} by
\sgg{a} cross-entropy loss and a \sgg{conventional} distance-based classifier, along
with an image retrieval view, to employ a re-ranking process to calibrate \sgg{a target} distance matrix by discovering the $k$-reciprocal neighbours within the task. 
Assuming the pre-trained \sgg{representation} is biased towards the source, we construct a non-linear subspace to \sgg{minimise} task-irrelevant features \sgg{therewithin} while keep more \sgg{transferrable} discriminative information by a hyperbolic tangent transformation. 
The calibrated distance in this \sgg{target-aware} non-linear subspace is complementary to that in the pre-trained \sgg{representation}.
To \sgg{impose such} distance calibration information \sgg{onto the pre-trained representation}, a Kullback-Leibler divergence loss is employed to gradually \sgg{guide} the model \sgg{towards} the calibrated distance-based distribution. 
Extensive \sgg{evaluations} on eight target
domains show that this \sgg{target ranking} calibration process can \sgg{improve conventional}
distance-based classifiers in \sgg{few-shot learning}.
\end{abstract}

\section{Introduction}
\label{sec:intro}

%
Few-Shot Learning (FSL) \sgg{promises to allow a} machine to learn novel concepts from limited experience, \ie few novel target data \sgg{and data-rich} source data. 
Typically, the defaulted FSL assumes that the source and target data
\sgg{is} in the same domain, but belong to different classes.
\sgg{In practice, FSL is required to generalise to different target domains}. 
\sgg{Cross-Domain Few-Shot Learning} (CD-FSL)
~\cite{tseng2020cross_fwt,guo2020broader,liang2021boosting_cdfsl,fu2021metamixup,hu2022switch}
has been studied \sgg{more recently}. \sgg{In CD-FSL, the target data not only has a different label space but also are from a different domain to the source data.}


%

\begin{figure}[t]
  \centering
   \includegraphics[width=1.0\linewidth]{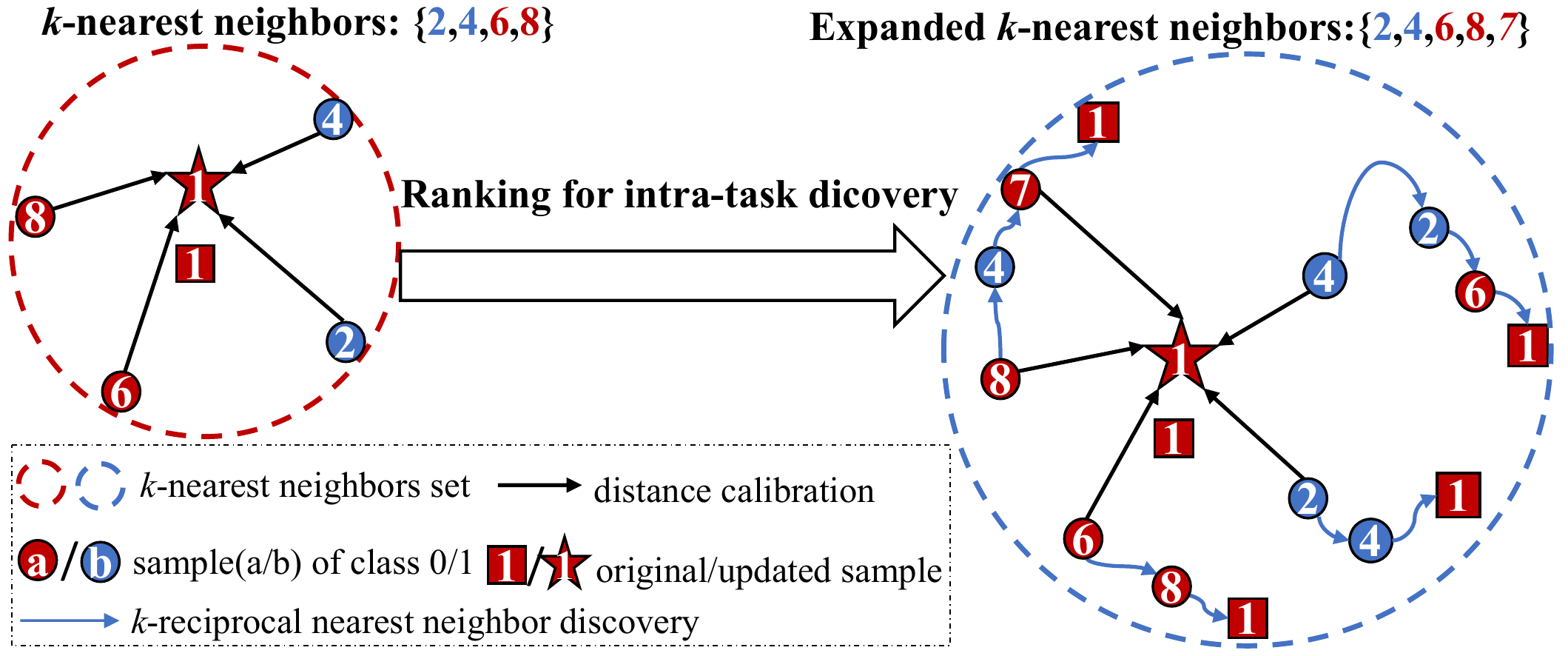}
   \vspace{-0.2in}
   \caption{\small{}\camera{\textbf{An illustration of ranking distance calibration
       process in a FSL task}. The idea is to \sgg{first} discover
     \sgg{likely} positive samples (\eg sample 7) for each instance
     (\eg sample 1) and then \sgg{to} calibrate their pairwise distances. \sgg{This is} achieved by mining the
     reciprocal ranking relations \sgg{for each instance retrieval
       task in the target domain so} to expand the $k$-nearest neighbours set.} \label{fig:motivation} }
  \vspace{-0.1in}
\end{figure}

It is nontrivial to directly extend the general FSL \sgg{approach} to address the CD-FSL challenges. 
In fact, 
many promising FSL
methods~\cite{snell2017prototypical,finn2017model,sung2018learning,satorras2018fewgnn}
performed poorly \sgg{in CD-FSL}~\cite{guo2020broader,tseng2020cross_fwt}.
The \sgg{central} idea of these general FSL methods is \sgg{to
  transfer and generalise the} visual representations learned \sgg{from source data to target data}.
However, \sgg{the significant} visual domain gap between \sgg{the}
source and target data \sgg{in CD-FSL} makes it fundamentally
  difficulty \sgg{to learn a shared visual representation across} different domains.



%

\sgg{A few recent CD-FSL
  studies}~\cite{tseng2020cross_fwt,liu2020urt,liang2021boosting_cdfsl,wang2021cross_ata}
try to learn a generalisable feature extractor to improve model
transferability, which is \sgg{a popular idea} in domain generalisation and
domain
adaptation~\cite{volpi2018generalizing_dataaug,NEURIPS2020_adver_dataaug,
  zhou2021dg_mixstyle,li2021simple,Wu_2021_ICCV} where the source and target
domains share the same label space. 
Empirically, this \sgg{approach} shows some improvement on CD-FSL but
\sgg{it does not model any visual and label
  characteristics} of the target domain \sgg{and more importantly
  their cross-domain impact on the pre-trained source domain
  representation. We argue this cross-domain mapping between 
  the source domain representation and its interpretion in the context of
  the target domain data characteristics is essential for effective
  CD-FSL.}
\sgg{From a related perspective, other CD-FSL studies have considered
  fine-tuning the source domain feature representation from augmenting
  additional support data in the target domain}, \eg \sgg{either explicitly} augmenting
the support data by adversarial training~\cite{wang2021cross_ata}
\sgg{and image} transformations~\cite{guo2020broader}, \sgg{or
  implicitly} augmenting the support data by training an
auto-encoder~\cite{liang2021boosting_cdfsl}. 
\sgg{However, these methods for CD-FSL are straightforward
  data-augmentation methods for increasing training data in target domain model
  fine-tuning, without considering how to quantify cross-domain relevance of the pre-trained
  source domain representation.}

\sgg{In this work, we consider an alternative approach with a new
  perspective to treat cross-domain few-shot learning as an image
  retrieval task. We wish to optimise model adaptation by leveraging
  target domain retrieval task context, that is, not only the labelled
  support data but also the unlabelled query data.}
\sgg{To that end, we use a generic representation} pre-trained by
\sgg{a cross entropy loss} and a simple distance-based classifier as a
baseline, then employ a $k$-reciprocal neighbour discovery (as in
Fig.~\ref{fig:motivation}) and encoding process to calibrate pairwise
distances \sgg{between each unlabelled query image and its likely
  matches}. 
Our \sgg{idea is both orthogonal and complementary} to other
generalisable model learning
methods~\cite{guo2020broader,liu2020feature,liang2021boosting_cdfsl}.
It can be flexibly used in \sgg{either fine-tuning or without fine-tuning
based model learning.}

Generally, the distance matrix for CD-FSL task contains many incorrect
results as this distance is \sgg{built} on a \sgg{potentially} biased
pre-trained \sgg{source domain representational} space. To calibrate
\sgg{this} distance matrix \sgg{towards the target domain so to reduce
  its bias to the source domain}, we \sgg{explore} the re-ranking concept
\sgg{in the target domain by considering CD-FSL optimisation as
  re-ranking in a retrieval task given few-shots as anchor points}.
As in Fig.~\ref{fig:motivation}, \sgg{re-ranking} first computes a
$k$-nearest neighbour ranking list. \sgg{This is} further expanded by
discovering the $k$-reciprocal nearest neighbours \sgg{in the target domain}. The expanded ranking
list is used for re-computing a Jaccard distance to measure the
difference between the original ranking list and the expanded ranking
list, achieving a more robust and accurate distance matrix. 
\sgg{
 {Critically, a pre-trained representation from source domain is
  biased and poor for generalisation cross-domains in CD-FSL. The reason is that conventional
  FSL methods assume implicitly linear transformations mostly between the source and
  target data as they are sampled from the same domain.}
  This becomes invalid in CD-FSL with mostly nonlinear
  transformations across source and target domains. To address this
  problem, we propose a task-adaptive subspace mapping to
  minimise transferring task-irrelevant representational information
  from the source domain. In particular,
  we explore a hyperbolic tangent function to
  project the source domain representation to a non-linear
  space}. Compared to the linear Euclidean space, this non-linear space
\sgg{performs a dimensionality reduction to optimise the retention of
  transferrable information from the source to the target domain.
Moreover, we explore the idea of re-ranking to
calibrate and align two distance matrices in two representational spaces
between the original pre-trained source domain linear space and the new
non-linear subspace. The calibrated matrices are combined to construct a
single distance matrix for the target domain in CD-FSL. We call this Ranking Distance Calibration (RDC).}
\sgg{To impose the above distance calibration into the representational
  space transform, we approximate the distance matrices by their
  corresponding distributions, and then a Kullback-Leibler (KL)
  divergence loss function is optimised for iteratively mapping the
  original distance distribution from the source domain towards the
  calibrated space. This provides an additional RDC Fine-Tuning
  (RDC-FT) model optimisation.}

\sgg{Our \textbf{contributions} from this work are three-fold:}
  \sgg{(1) To transform the biased distance matrix in the source
      domain representational space towards the target domain in CD-FSL, we
      use a re-ranking method to re-compute a Jaccard distance for
      distance calibration by discovering the reciprocal nearest neighbours
      within the task. We call this Ranking Distance Calibration.
      (2) We propose a non-linear subspace to shadow the
      pre-trained source domain representational space. This is
      designed to model any inherent non-linear transform in CD-FSL and
      used to facilitate the distance calibration process between the
      source and target domains. By modelling explicitly this
      nonlinearity, we formulate a more robust and generalisable
      Ranking Distance Calibration (RDC) model for CD-FSL.
      (3) We further impose RDC as a constraint to the model optimisation process. This
      is achieved by a RDC with Fine-Tuning (RDC-FT) for iteratively
      mapping the original source domain distance distribution to a calibrated
      target domain distance distribution for a more stable and improved
      CD-FSL. 

We evaluated the proposed RDC and RDC-FT methods \sgg{for CD-FSL on} eight
target domains. \sgg{The} results show that RDC can \sgg{improve
  notably} the \sgg{conventional} distance-based classifier,  {and RDC-FT can improve the representation for target domain} to achieve
competitive \sgg{or} better performance than the \sgg{state-of-the-art
  CD-FSL models}. 
}

\section{Related Work}
\label{sec:related}
\keypoint{Few-shot learning.}
 {
The approaches for general FSL can be broadly divided into two categories: optimisation-based methods~\cite{finn2017model,ravi2016optimization,oh2021boil} which learn a generalisable model initialisation and then adapt the model on a novel task with limited labelled data, and metric learning methods~\cite{snell2017prototypical,sung2018learning,zhang2020deepemd,li2021plain} that meta-learn a discriminative embedding space where the sample in novel task can be well-classified by a common or learned distance metric. Recently, some studies~\cite{chen2019closer,tian2020rethinking,li2020few_self} 
show that a simple pre-training method followed by a fine-tuning stage can achieve competitive or better performance than the metric learning methods.
This observation also seems to be true in CD-FSL~\cite{guo2020broader}.
}

\keypoint{Cross-domain few-shot learning.}
 {
The problem of CD-FSL was preliminarily studied in FSL~\cite{chen2019closer,tian2020rethinking,pan2021mfl}, then~\cite{tseng2020cross_fwt,guo2020broader} expanded this setup and proposed two benchmarks to train a model on a single source domain and then generalise it to other domains.
Some CD-FSL studies~\cite{tseng2020cross_fwt,wang2021cross_ata,liang2021boosting_cdfsl} focus on learning a generalisable model \camera{from the source domain} by explicit or implicit data augmentation.
These approaches improve the model generalisation ability but easily result in ambiguous optimisation result since they ignore the adaption process for the target domain.
\camera{To address the domain shift problem \new{in a feature representation}, 
CHEF~\cite{adler2020cross} uses a fusion strategy for feature ensemble
whilst ConFeSS~\cite{das2022confess} learns a mask to select relevant
\new{features in} the target domain.}
Another methods~\cite{phoo2021STARTUP,das2021distractor_cdfsl,fu2021metamixup} target to the adaptation on the target domain by leveraging additional unlabelled data~\cite{phoo2021STARTUP}, labelled data~\cite{fu2021metamixup} or the base data~\cite{das2021distractor_cdfsl}.
In practice, the increased data can help model adaptation on the target domain but these information are not easy to obtain.
In this work, we address the CD-FSL problem with an image retrieval view and mine the intra-task information to guide a ranking distance calibration process. 
}

\begin{figure*}[ht]
  \centering
  \vspace{-0.05in}
   \includegraphics[width=0.95\linewidth]{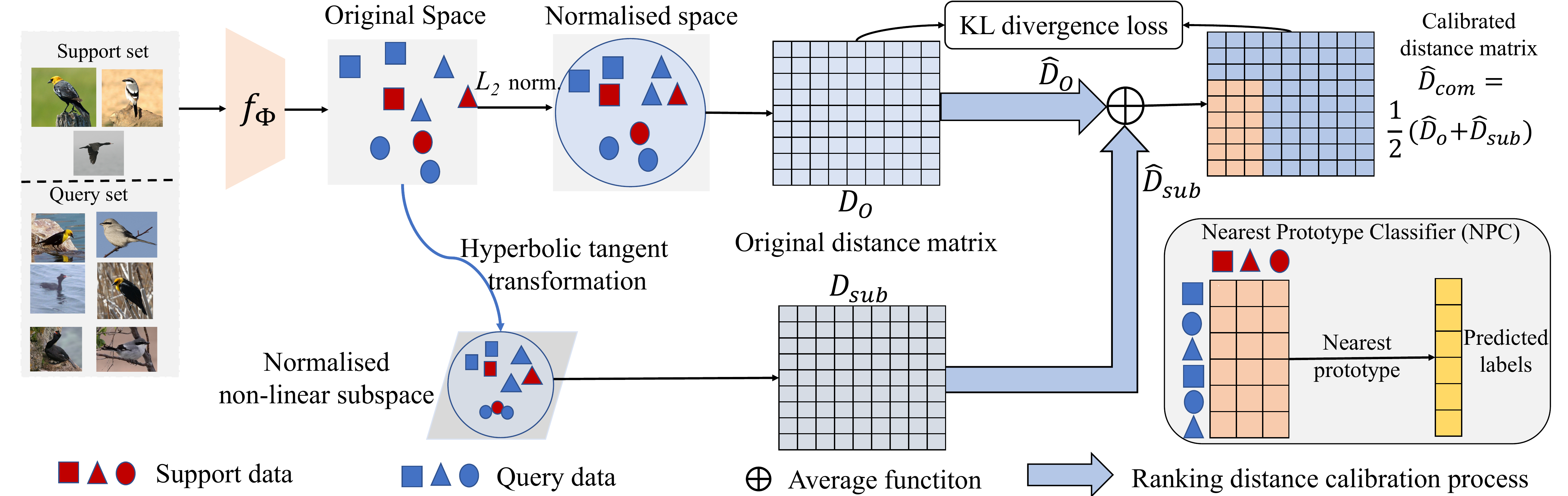}
  \vspace{-0.05in}
   \caption{ {\small{}\textbf{An overview of the proposed ranking distance calibration pipeline.} $f_{\Phi}$ is the feature extractor pre-trained on the source dataset with a standard cross entropy loss. Our \textbf{RDC} method contains two parts: 1) a ranking process on the original space to calibrate the $D_o$ to $\hat{D}_o$, and 2) a ranking process on a non-linear subspace to calibrate the $D_{sub}$ to $\hat{D}_{sub}$.  The proposed \textbf{RDC-FT} method uses a KL loss between $D_o$ and $\hat{D}_{com}$ to fine-tune $f_{\Phi}$. The NPC is used to classify the query data according to the pairwise distances, \ie the calibrated distances $\hat{D}_{com}$ by RDC and the Euclidean distances on the embeddings fine-tuned with RDC-FT, respectively.}}
\vspace{-0.1in}
   \label{fig:framework}
\end{figure*}

\keypoint{Ranking in image retrieval.}
\camera{Image Retrieval (IR) is a classical \new{computer} vision task
  that aims to search \new{in} unlabelled gallery data \new{in order} to find \new{those} images
  that are most \new{similar} to \new{a} probe image. \new{In} IR,
\new{a} ranking
method~\cite{he2004manifold_rank,huang2015cross_retrivel,liu2013pop_rank,loy2013person_rank}
\new{computes} a \new{rank} list \new{by} a distance
metric. \new{Moreover, different}
re-ranking ideas~\cite{zhong2017rerank,sarfraz2018posererank,wang2020high} were
proposed as a post-processing \new{mechanism} to improve the initial ranking
result. For instance, \new{a model} \cite{zhong2017rerank} uses \new{the} concept of
$k$-reciprocal nearest neighbours~\cite{qin2011hello_kreciprocal} to
explore more hard positive samples. \new{The} enlarged $k$-nearest
neighbours are \new{then} used to recompute a Jaccard distance as an auxiliary
distance matrix. 
This idea \new{was further adopted by others for
  IR}~\cite{liu2018adaptivererank,sarfraz2018posererank,wang2020high}.}  

\camera{
In this work, we \new{explore the concept of re-ranking}
\new{in few-shot learning. Although related to the work
  of~\cite{xi2021_reranking}, our approach differs. The idea
  of~\cite{xi2021_reranking} is to improve subgraph similarity by a
graph view {\em solely in a pre-trained space}. In contrast,} our idea is to calibrate
\new{a} similarity distance \new{{\em both}} in a pre-trained space and a task-adaptive subspace. To
optimise \new{a} feature \new{representation}, we use \new{this} calibrated distance to guide
cross-domain knowledge transfer with a Kullback-Leibler divergence
loss. \new{The model of~\cite{xi2021_reranking} does not learn cross-domain
knowledge in re-ranking.} It uses a MLP to meta-learn a
subgragh refiner \new{{\em in a single domain}} with a cross entropy loss. 
}

\section{Methodology \label{sec:method}}
\noindent\textbf{Problem formulation.}
We start by defining a general FSL problem: 
given a source dataset $\mathcal{D}_s$ and a target dataset $\mathcal{D}_t$, where the classes in $\mathcal{D}_s$ and $\mathcal{D}_t$ are disjoint. FSL aims to address the $C$-way $K$-shot classification task $\mathcal{T}$ in $\mathcal{D}_t$ by leveraging the limited data in $\mathcal{T}$ and the prior knowledge learned from $D_s$ containing lots of labelled images. In specific, a FSL task $\mathcal{T}$ contains a labelled support set $\mathcal{T}_s=\{\mathbf{I}_{i}, y_{i}\}_{i=1}^{C\times K}$ and an unlabelled query set $\mathcal{T}_{q}=\{\mathbf{I}_{j}\}_{j=1}^{C\times Q}$, where the images in $\mathcal{T}_s$ and $\mathcal{T}_q$ are both from the same $C$ classes and $K$/$Q$ denote the number of images per class in $\mathcal{T}_s$/$\mathcal{T}_q$.
The goal of FSL is to recognise the unlabelled query set $\mathcal{T}_q$ when $K$ is small.
\camera{\new{In} the CD-FSL setup, $\mathcal{D}_s$ and $\mathcal{D}_t$
  are from different domains, \eg $\mathcal{D}_s$ contains \new{many}
  natural \new{photos} whilst the images in $\mathcal{D}_t$ are
  \new{captured by remote sensing sensors}.} 

\keypoint{Nearest prototype classifier.} 
We \camera{firstly} define the prototype classifier \camera{used in this \new{work}.} 
Given a feature extractor $f$, we can extract the embedding ${x}=f(\mathbf{I})$ for image $\mathbf{I}$ in a FSL task $\mathcal{T}$. Nearest Prototype Classifier (NPC) first computes the prototypes $ {\hat{X}}=\{ {\hat{x}}_{0},\cdots, {\hat{x}}_{C-1}\}$ for the $C$ classes, where the prototype for class $c$ is:
\vspace{-0.05in}
\begin{equation}
    \hat{ {x}}_c=\frac{1}{K}\sum_{i=1}^{K} {x}_i,\quad \mathrm{where} \;  {x}_i \in \mathcal{T}_s, y_i=c.
\end{equation}
\vspace{-0.05in}
With the prototypes, the labels for $x_j$ in $\mathcal{T}_q$ is assigned by:
\begin{equation}
    y({ {x}}_j)=\mathrm{argmin}\,_{c\in\{0,\cdots,C-1\}}\;d( {x}_j,\hat{ {x}}_c),
    \label{eq:classification}
\end{equation}
where $d(\cdot)$ is a distance metric, \eg Euclidean distance in this work, and $d( {x}_j,\hat{ {x}}_c)$ is the distance between $ {x}_j$ and $\hat{ {x}}_c$. 

\keypoint{Overview.}
The key insight of this paper is to formulate the FSL as the Image Retrieval (IR) task by sharing the same angle with~\cite{triantafillou2017fewretrivel}.
In~\cite{triantafillou2017fewretrivel}, the authors propose to optimise the mean average precision for FSL.
Furthermore, our view of ``FSL as IR'' also emphasises the importance of maximally leveraging all the available information in this low-data regime whilst concerns on the calibration of pairwise distances in FSL. In particular, this work follows this view for FSL and consider each sample in FSL as the probe data in IR and treat the whole FSL data as the gallery data.
To this end, we propose a ranking distance calibration process for CD-FSL, and our key methodology is to repurpose the re-ranking to find the relevant images from the FSL task for a given image.  We overview the proposed method in Fig.~\ref{fig:framework}.

\subsection{Ranking Distance Calibration (RDC)}
\label{subsec:rerank}
\noindent\textbf{Motivation.} 
Previous works~\cite{zhong2017rerank,qin2011hello_kreciprocal} have suggested that discovering the $k$-reciprocal nearest neighbours within the gallery data can benefit the re-ranking result for image retrieval.
This observation encourages us, when considering few-shot as image retrieval, to reuse this $k$-reciprocal nearest neighbour discovery process in~\cite{zhong2017rerank} to calibrate the pairwise distances within FSL task.
\camera{\new{Fig.~\ref{fig:motivation} gives an illustration: The}
  $k$-reciprocal nearest neighbour discovery process finds more
  hard-positive samples for a given query sample.} 
\camera{These \new{hard-positive} samples are then used to
  \new{update each} pairwise distance by $k$-reciprocal encoding and
  \new{to} further estimate a better distance by query
  expansion. \new{These} pairwise distances in \new{a} FSL task
  \new{are} calibrated and \new{represented in} a new distance matrix.}

\new{Let us now describe} the re-ranking process \new{in} our ranking
distance calibration. \camera{\new{We further describe in details} the Jaccard
distance computing process in the supplementary material}. For a FSL
  task, we start by computing an original pairwise Euclidean distance
  matrix: 
\begin{equation}
D_o=
\begin{bmatrix}
  d_o(1,1)&\cdots&d_o(1,n) \\
  \vdots &\ddots   &\vdots \\
  d_o(n,1)&\cdots  &d_o(n,n) 
\end{bmatrix}, \; \mathrm{where} \; n=C\times (K+Q).
\label{eq:matrix}
\end{equation}
$d_o(i,j)$ is the Euclidean distance between $\left \| x_i \right \|_2$ and $\left \| x_j \right \|_2$, and $\left \| \cdot \right \|_2$ represents the $l_2$ normalisation.
Referring to $D_o$, we can obtain the $k$-nearest neighbours set $ {R}=[ {R}_1(k), {R}_2(k),..., {R}_n(k)]$, and $R_i(k)$ is the $k$-nearest neighbours of $x_i$.
The re-ranking idea~\cite{zhong2017rerank} is to expand the $R_i(k)$ by discovering more hard-positive samples for $x_i$.
The expand process for $R_i(k)$ is guided by a $k$-reciprocal nearest neighbours algorithm~\cite{qin2011hello_kreciprocal} and the expanded ranking list $\hat{R}_i(k)$ is used to estimate a calibrated distance matrix.

\keypoint{\textbf{$k$}-reciprocal discovery and encoding.}
The principle of $k$-reciprocal nearest neighbour discovery is that if $x_g$ is in $R_i(k)$, then $x_i$ should also occur in $R_g(k)$~\cite{qin2011hello_kreciprocal}.
\new{Given} this assumption, \camera{we \new{adopt the method
    of}~\cite{zhong2017rerank} \new{for computing} the expanded
  $\hat{R}_i(k)$, \new{defined as}:}
\begin{equation}
\begin{aligned}
&\hat{ {R}}_i(k) \leftarrow {R}_i(k) \cup \mathcal{R}_g((1/2)k) \\
& s.t. \left | {R}_i(k) \cap  {R}_g((1/2)k)   \right |  \ge (2/3) \left |  {R}_g((1/2)k) \right|,
\end{aligned}
\label{eq:distance}
\end{equation}
where $x_g$ is the sample in $\mathcal{R}_i(k)$ and $|\cdot|$ is the number of neighbours.
\camera{In particular, to \new{leverage effectively} the labelled
  support data in FSL, we remove the labelled distractors from
  neighbours and further expand $\hat{ {R}}_i(k)$ by uniting $\hat{
    {R}}_s(k)$, where $x_s$ and $x_i$ are both in the support set as
  well \new{as from the} same class. Finally, the expanded $k$-nearest
  neighbours set \new{is}
  $\hat{R}=[\hat{R}_1(k),\cdots,\hat{R}_n(k)]$.}
To assign larger weights to closer neighbours while smaller weights to farther ones, the $\hat{R}_i(k)$ is further used to encode the $\mathbf{d}_o(i,:)=[d_o(i,g_1),\ldots,d_o(i,g_n)]$ into a vector $\mathcal{V}_i=[\mathcal{V}_{i,g_1}, \mathcal{V}_{i,g_2},...,\mathcal{V}_{i,g_n}]$, where $\mathcal{V}_{i,g_i}$is defined as the Gaussian kernel of the pairwise distance as:
\begin{equation}
\mathcal{V}_{i,g_q}=\begin{cases}
  e^{-d_o(x_i,x_{g_q})}& \text{ if } g_q\in\hat{\mathcal{R}}_i(k) \\
  0 & \text{otherwise.} 
\end{cases}
\label{eq:re-weight}
\end{equation}
After that, a \camera{query expansion strategy~\cite{chum2007total,qin2011hello_kreciprocal}} is employed to integrate $k_{2}$ most-likely samples to update the feature of $x_{i}$ by: $\mathcal{V}_{i}=(1/|\hat{R}_i(k_2)|){\textstyle \sum_{g_q\in \hat{R}_i(k_2)}\mathcal{V}_{g_q}},$ where $k_{2}<k$.

\keypoint{Jaccard distance.}
 {Referring to ~\cite{bai2016sparse,zhong2017rerank}, the expanded ranked list $\hat{R}$ is used as contextual knowledge to compute a Jaccard distance matrix} $D_{J}=\{d_J(i,g_q) | q \in [1,n]\}$ by:
\begin{equation}
d_J(i,g_q)=1-\frac{|\hat{R}_i(k) \cap \hat{R}_{g_q}(k) |}{| \hat{R}_i(k) \cup \hat{R}_{g_q}(k)|}.
\label{eq:jaccard}
\end{equation}
Following the re-weighting method in~\cite{zhong2017rerank}, the number of candidates in the intersection and union set can be calculated as $|\hat{ R}_{g_q}(k) \cap \hat{R}_i(k)|=||\mathrm{min}(\mathcal{V}_i, \mathcal{V}_{g_q})||_1$ and $|\hat{R}_{g_q}(k) \cup \hat{\mathcal{R}}_i(k)|=||\mathrm{max}(\mathcal{V}_i, \mathcal{V}_{g_q})||_1$,
where $\mathrm{min}$ and $\mathrm{max}$ operate the element-based minimisation and maximisation for two input vectors, and $||\cdot||_1$ is $l_1$ norm. Then the Jaccard distance in Eq.(\ref{eq:jaccard}) can be re-formulated as:
\begin{equation}
d_J(i,g_q)=1-\frac{ {\textstyle \sum_{j=1}^{n}}\mathrm{min}(\mathcal{V}_{i,g_j}, \mathcal{V}_{g_q,g_j}) }{{\textstyle \sum_{j=1}^{n}}\mathrm{max}(\mathcal{V}_{i,g_j}, \mathcal{V}_{g_q,g_j})}.
\label{eq:jaccard_v}
\end{equation}

\keypoint{Distance calibration.}
\camera{The Jaccard distance exploits the contextual information to compute a relative distance in context within $k$-reciprocal neighbours. The original distance is an absolute distance in pre-trained Euclidean space. Therefore, combing $D_{J}$ and $D_{o}$ makes the distance more discriminative among neighbours by a weighting strategy:}
\begin{equation}
    \hat{D}_o= \lambda D_o + (1-\lambda)D_{J},
    \label{eq:calibrate}
\end{equation}
where $\lambda$ is a trade-off scalar to balance the two matrices.

\subsection{RDC in Task-adaptive Non-linear Subspace}
\label{subsec:non-linear}
To further bridge the domain gap, we propose further improving the RDC in a non-linear subspace. We particular tailor a  discriminative subspace to help calibrate the ranking in our  CD-FSL task.
The subspace is built upon the Principal Component Analysis (PCA) to extract crucial features from the original space.
In specific, given the feature representations $ {X}\in \Re^{n\times m}$ of a target FSL task, we have
\begin{equation}
     {X}_{sub}= {X} {P}, \;\;  \mathrm{where}\;\mathcal{P}\in \Re^{m\times p}, {X}_{sub}\in \Re^{n\times p}.
    \label{eq:sub}
\end{equation}
$\mathcal{P}$ is a transformation matrix mapping the feature with $m$ dimensions to a reduced feature with $p$ dimensions. 

\keypoint{Hyperbolic tangent transformation.}
 {
Generally, the PCA method can be directly used on the original embedding space. However, the original representation $ {X}$ is scattered due to the biased and less-discriminative embedding; thus the dimensional reduction easily causes the information loss problem. To remit this issue, we consider to transform the original representations to a compact and representative non-linear space.
By using the idea of kernels,
we use a hyperbolic tangent function to construct a task-adaptive non-linear subspace.
Our non-linear PCA method first computes a feature-toward kernel function by:
}
\begin{equation}
    \mathcal{K}=\mathrm{tanh}( {X}^T {X}), \;\mathcal{K}\in \Re^{m\times m}.
    \label{eq:kernel}
\end{equation}
Then we use Singular Value Decomposition (SVD) to compute the eigenvalues $U$ of $\mathcal{K}$ and select the most $p$-relevant eigenvalues $U^p$, formulating the transformation matrix $\mathcal{P}=U^p$. To this end, a task-adaptive non-linear subspace $ {X}_{sub}$ is construct by Eq.(\ref{eq:sub}) and Eq.(\ref{eq:kernel}).

\keypoint{Complementary distance calibration.}
Our distance calibration process \camera{is space-agnostic} and can be applied in the original linear space (in Sec.~\ref{subsec:rerank}) and a non-linear subspace (in Sec.~\ref{subsec:non-linear}).
The original space has higher dimensions consist of full information but also disturbed by noisy task-irrelevant features, while the non-linear subspace reduce some task-irrelevant signal but loss some information. Our RDC method co-leverages the calibrated distances in the two spaces 
to capture a robust and complementary distance matrix $\hat{D}_{com} = 0.5(\hat{D}_{o}+\hat{D}_{sub})$.
\new{How to compute RDC in two spaces} is \new{described in lines} 4-8 of Alg.~\ref{algorithm}.

\keypoint{Remark.}
 {
As the hyperbolic non-linear space has larger capability than Euclidean space~\cite{khrulkov2020hyperbolic,fang2021kernel_hyperbolic,yan2021unsupervised_hyperbolic}, it can alleviate the information loss caused by the dimensionality reduction.
Hence, we use a hyperbolic tangent transformation to map the source domain linear space to a non-linear space.
We note that the subspace learning has been preliminary explored in FSL work~\cite{yoon2019tapnet,simon2020adaptive} to learn task-adaptive or class-adaptive subspace.
Critically, our subspace construction method \camera{differs} from~\cite{yoon2019tapnet,simon2020adaptive} \camera{as} our method does not need the sophisticated episode training process.}

\subsection{Fine-tuning with RDC}
\label{sec:fine_tune}
As $\hat{D}_{com}$ provides a more robust and discriminative distance matrix, it is natural to ask whether this type of calibration knowledge can be used to optimise the feature extractor. To achieve this, we fine-tune the feature extractor by iteratively mapping the original distance distribution to the calibrated distance distribution, formulating \camera{a} RDC with Fine-Tuning (RDC-FT) method as in Alg.~\ref{algorithm}.

\begin{algorithm}[t]
\caption{{Ranking Distance Calibration with Fine-Tuning (RDC-FT)}}
\label{algorithm}
\KwData{pre-trained feature extractor $f_{\Phi}$; support set $\mathcal{T}_s$; query set $\mathcal{T}_q$; RDC: $k_1$, $k_2$, $\lambda$; Fine-tune: epochs $T$, learning rate $\beta$, $\tau, \alpha$.}
\KwResult{Fine-tuned feature extractor $f_{\hat{\Phi}}$.}
   \tcc{ RDC-FT: optimise $f_{{\Phi}}$ by RDC}
    Initialise $\hat{\Phi} =\Phi $ \;  
    \For{$t$ in $T$}
    {
    Extract embeddings for $\mathcal{T}_s$ and $\mathcal{T}_q$ by $f_{\hat{\Phi}}$\;
    \tcc{ RDC: distance calibration}
    Compute original distances $D_o$ by Eq.(\ref{eq:matrix}) \;
    Compute calibrated distances $\hat{D}_{o}$ by Eq.(\ref{eq:calibrate}) \;
    Construct a non-linear subspace by Eq.(\ref{eq:sub}, \ref{eq:kernel}) \;
    Calibrate distance in subspace {$\hat{D}_{sub}$} by Eq.(\ref{eq:calibrate}) \;
    Compute $\hat{D}_{com}=0.5*(\hat{D}_o+\hat{D}_{sub})$\;
    \tcc{ FT: optimise $f_{\hat{\Phi}}$ with $\hat{D}_{com}$}
    Get the softened distribution $\mathbf{p}^{{D}_{o}}(\tau)$/$\mathbf{p}^{\hat{D}_{com}}(\tau)$ \;
    Compute KL loss by Eq.(\ref{eq:klloss})\;
    $\hat{\Phi} \leftarrow {\hat{\Phi}}-\beta \bigtriangledown_{\hat{\Phi}}L_{KL}(\mathbf{p}^{D_o}(\tau),\mathbf{p}^{\hat{D}_{com}}(\tau))$ \; 
    
    }
\end{algorithm}

\keypoint{Expanded $k$-reciprocal list as attention. } 
 {
As in Eq.~(\ref{eq:jaccard}), the expanded ranking list $\hat{R}(k)$ is used to re-compute the pairwise distances. The calibrated pairwise distances in $\hat{R}(k)$ are more robust than these not in $\hat{R}(k)$. Thus the $\hat{R}(k)$ can naturally be used as an attention mask $\mathcal{M}$. In particular, a $\mathcal{M}$ is computed by  
\begin{equation}
\mathcal{M}_i^{g_q}=\begin{cases}
  1+\alpha& \text{ if } g_q\in\hat{\mathcal{R}}_i(k) \\
  1 & \text{otherwise,} 
\end{cases}
\label{eq:attention}
\end{equation}
where $\alpha$ is an attention scalar. During the fine-tuning process, the $\mathcal{M}$ is used to re-weight the distance matrices $D_o$ and $\hat{D}_{com}$ as $\mathcal{M}\cdot D_o$ and $\mathcal{M} \cdot \hat{D}_{com}$, respectively.
}
\begin{table*}[t]
\centering
\scalebox{0.78}{
\begin{tabular}{lccccccccc}
\hline
\multirow{2}{*}{Method} &\multicolumn{9}{c}{{5-way 1-shot}} \\
\cline{2-10}
&CUB &Cars &Places &Plantae &\small{}CropDisease &EuroSAT &ISIC &ChestX &Ave. \\
\hline
ProtoNet~\cite{snell2017prototypical}
&38.66\small{}$\pm$0.4 &31.34\small{}$\pm$0.3 &47.89\small{}$\pm$0.5 &31.75\small{}$\pm$0.4  &51.22\small{}$\pm$0.5 &52.93\small{}$\pm$0.5 &29.20\small{}$\pm$0.3 &{21.57\small{}$\pm$0.2} &38.07 \\
NPC
&38.47\small{}$\pm$0.4 &33.27\small{}$\pm$0.3 &40.84\small{}$\pm$0.4 &36.77\small{}$\pm$0.4 &64.76\small{}$\pm$0.5 &51.45\small{}$\pm$0.5 &29.46\small{}$\pm$0.3 &22.74\small{}$\pm$0.2 &39.72 \\
NPC+$l_2$ norm
&43.67\small{}$\pm$0.4 &35.76\small{}$\pm$0.4 &48.67\small{}$\pm$0.4 &39.15\small{}$\pm$0.5  &66.62\small{}$\pm$0.5 &60.85\small{}$\pm$0.5 &31.52\small{}$\pm$0.3 &\textbf{22.87\small{}$\pm$0.2} &43.64 \\
RDC (ours)
&\textbf{48.68\small{}$\pm$0.5} &\textbf{38.26\small{}$\pm$0.5} &\textbf{59.53\small{}$\pm$0.5} &\textbf{42.29\small{}$\pm$0.5}  &\textbf{79.72\small{}$\pm$0.5} &\textbf{65.58\small{}$\pm$0.5} &\textbf{32.33\small{}$\pm$0.3} &{22.77\small{}$\pm$0.2}  &\textbf{48.65} \\
\hline
\hline
\multirow{2}{*}{Method} &\multicolumn{9}{c}{{5-way 5-shot}} \\
\cline{2-10}
&CUB &Cars &Places &Plantae &\small{}CropDisease &EuroSAT &ISIC &ChestX &Ave. \\
\hline
ProtoNet~\cite{snell2017prototypical}
&{57.55\small{}$\pm$0.4} &43.98\small{}$\pm$0.4  &68.05\small{}$\pm$0.4 &46.18\small{}$\pm$0.4  &79.98\small{}$\pm$0.3 &{75.36\small{}$\pm$0.4} &{39.98\small{}$\pm$0.3} &{24.19\small{}$\pm$0.2}  &54.41 \\
NPC
&60.48\small{}$\pm$0.4 &51.16\small{}$\pm$0.4 &67.74\small{}$\pm$0.4 &53.34\small{}$\pm$0.4  &86.95\small{}$\pm$0.3 &74.27\small{}$\pm$0.4 &39.26\small{}$\pm$0.3 &26.17\small{}$\pm$0.2 &57.42 \\
NPC+$l_2$ norm
&63.23\small{}$\pm$0.4 &{51.92\small{}$\pm$0.4} &69.95\small{}$\pm$0.4 &55.76\small{}$\pm$0.4  &87.76\small{}$\pm$0.3 &76.29\small{}$\pm$0.4 &41.08\small{}$\pm$0.3 &\textbf{26.31\small{}$\pm$0.2} &59.03\\
RDC (ours)
&\textbf{64.36\small{}$\pm$0.4} &\textbf{52.15\small{}$\pm$0.4} &\textbf{73.24\small{}$\pm$0.4} &\textbf{57.50\small{}$\pm$0.4}  &\textbf{88.90\small{}$\pm$0.3} &\textbf{77.15\small{}$\pm$0.4} &\textbf{41.28\small{}$\pm$0.3} &25.91\small{}$\pm$0.2 &\textbf{60.06} \\
\hline
\end{tabular}
}
\vspace{-0.1in}
\caption{\small{} {\textbf{Comparisons with the baselines using NPC
      classifier \textit{w/o} fine-tuning.}  The classification
    accuracies on 8 datasets with \textbf{ResNet10} as the
    backbone. \camera{RDC exploits the labelled support data and the
      unlabelled query data \new{in a} FSL task.} \textbf{Bold:} The best scores.} \label{tab:baseline}}
\vspace{-0.1in}
\end{table*}

\keypoint{Choices of loss functions.}
To achieve the distance distribution alignment, Mean Squared Error (MSE) loss and Kullback-Leibler (KL) divergence loss are alternatives.
The MSE loss prefers to directly learn towards the target distance while KL divergence loss focuses on the distribution matching~\cite{kim2021comparing}. As KL loss learns this mapping process in a softening way, it is a better way to embed the calibration knowledge into the representations. \camera{Thus we use KL loss}:
\begin{equation}
    \mathcal{L}_{KL}(\mathbf{p}^{D_o}(\tau),\mathbf{p}^{\hat{D}_{com}}(\tau))=\tau^2\sum_{j}\mathbf{p}_j^{\hat{D}_{com}}(\tau) \mathrm{log}\frac{\mathbf{p}_j^{\hat{D}_{com}}(\tau)}{\mathbf{p}_j^{D_{o}}(\tau)},
    \label{eq:klloss}
\end{equation}
 {where $\tau$ is the temperature-scaling hyper-parameter, $\mathbf{p}_j^{D_{o}}(\tau)$ and $\mathbf{p}_j^{\hat{D}_{com}}(\tau)$
are the softened distributions of the re-weighted distances matrices $\mathcal{M}\cdot D_o$ and $\mathcal{M}\cdot \hat{D}_{com}$. Given a vector $\mathbf{d}(i,:)$ in the distance matrix $D$, the softened distribution $\mathbf{p}^{d(i,:)}(\tau)$ is denoted by $\mathbf{p}^{d(i,l)}(\tau)=\frac{\mathrm{exp}(d(i,l)/\tau)}{ \sum_{j=1}^{n}\mathrm{exp}(d(i,j)/\tau)}$, where $d(i,l)$ is the $l$-th value of $\mathbf{d}(i,:)$.
}

\begin{table*}[]
\centering
\scalebox{0.77}{
\begin{tabular}{lccccccccc}
\hline
\multirow{2}{*}{Method} &\multicolumn{9}{c}{{5-way 1-shot}} \\
\cline{2-10}
&CUB &Cars &Places &Plantae &\small{}CropDisease &EuroSAT &ISIC &ChestX &Ave. \\
\hline
GNN+FT{$^\dagger$}~\cite{tseng2020cross_fwt}
&45.50\small{}$\pm$0.5 &32.25\small{}$\pm$0.4  &53.44\small{}$\pm$0.5 &32.56\small{}$\pm$0.4  &60.74\small{}$\pm$0.5 &55.53\small{}$\pm$0.5 &30.22\small{}$\pm$0.3 &22.00\small{}$\pm$0.2 &41.53 \\
GNN+LRP{$^\dagger$}~\cite{sun2021explanationcdfsl}
&43.89\small{}$\pm$0.5 &31.46\small{}$\pm$0.4  &52.28\small{}$\pm$0.5 &33.20\small{}$\pm$0.4  &59.23\small{}$\pm$0.5 &54.99\small{}$\pm$0.5 &30.94\small{}$\pm$0.3 &22.11\small{}$\pm$0.2 &41.01 \\
TPN+ATA{$^{*}$}~\cite{wang2021cross_ata}
&{50.26\small{}$\pm$0.5} &34.18\small{}$\pm$0.4 &{57.03\small{}$\pm$0.5} &39.83\small{}$\pm$0.4 &77.82\small{}$\pm$0.5 &65.94\small{}$\pm$0.5 &{34.70\small{}$\pm$0.4} &21.67\small{}$\pm$0.2 &47.68 \\
Fine-tuning{$^\dagger$}~\cite{guo2020broader}
&43.53\small{}$\pm$0.4 &35.12\small{}$\pm$0.4 &50.57\small{}$\pm$0.4 &38.77\small{}$\pm$0.4 &73.43\small{}$\pm$0.5 &66.17\small{}$\pm$0.5 &34.60\small{}$\pm$0.3 &22.13\small{}$\pm$0.2 &45.54 \\
ConFT{$^{\ddagger}$}~\cite{das2021distractor_cdfsl}
&45.57\small{}$\pm$0.8 &\textbf{39.11\small{}$\pm$0.7} &49.97\small{}$\pm$0.8 &{43.09\small{}$\pm$0.8}  &69.71\small{}$\pm$0.9{\textcolor{red}{$^1$}} &64.79\small{}$\pm$0.8{\textcolor{red}{$^1$}} &34.47\small{}$\pm$0.6{\textcolor{red}{$^1$}} &\textbf{23.31{\small{}$\pm$0.4}}{\textcolor{red}{$^1$}} &46.25 \\ 
RDC-FT{$^{*\ddagger}$} (ours)
&\textbf{51.20\small{}$\pm$0.5} &\textbf{39.13\small{}$\pm$0.5} &\textbf{61.50\small{}$\pm$0.6} &\textbf{44.33{\small{}$\pm$0.6}} &\textbf{86.33\small{}$\pm$0.5} &\textbf{71.57\small{}$\pm$0.5} &\textbf{35.84\small{}$\pm$0.4} &22.27\small{}$\pm$0.2 &\textbf{51.53} \\
\hline
\hline
\multirow{2}{*}{Method} &\multicolumn{9}{c}{{5-way 5-shot}} \\
\cline{2-10}
&CUB &Cars &Places &Plantae &\small{}CropDisease &EuroSAT &ISIC &ChestX &Ave. \\
\hline
GNN+FT{$^\dagger$}~\cite{tseng2020cross_fwt}
&64.97\small{}$\pm$0.5 &46.19\small{}$\pm$0.4  &70.70\small{}$\pm$0.5 &49.66\small{}$\pm$0.4  &87.07\small{}$\pm$0.4 &78.02\small{}$\pm$0.4 &40.87\small{}$\pm$0.4 &24.28\small{}$\pm$0.2 &57.72 \\
GNN+LRP{$^\dagger$}~\cite{sun2021explanationcdfsl}
&62.86\small{}$\pm$0.5 &46.07\small{}$\pm$0.4  &71.38\small{}$\pm$0.5 &50.31\small{}$\pm$0.4  &86.15\small{}$\pm$0.4 &77.14\small{}$\pm$0.4 &44.14\small{}$\pm$0.4 &24.53\small{}$\pm$0.3 &57.82 \\
TPN+ATA*~\cite{wang2021cross_ata}
&65.31\small{}$\pm$0.4 &46.95\small{}$\pm$0.4 &{72.12\small{}$\pm$0.4} &55.08\small{}$\pm$0.4 &88.15\small{}$\pm$0.5 &79.47\small{}$\pm$0.3 &45.83\small{}$\pm$0.3 &23.60\small{}$\pm$0.2  &{59.57} \\
Fine-tuning{$^{\dagger \ddagger}$}~\cite{guo2020broader}
&63.76\small{}$\pm$0.4 &51.21\small{}$\pm$0.4 &70.68\small{}$\pm$0.4 &56.45\small{}$\pm$0.4  &89.84\small{}$\pm$0.3 &81.59\small{}$\pm$0.3 &49.51\small{}$\pm$0.3 &25.37\small{}$\pm$0.2  &61.06 \\
ConFT{$^{\ddagger}$}~\cite{das2021distractor_cdfsl}
&\textbf{70.53\small{}$\pm$0.7} &\textbf{61.53\small{}$\pm$0.7} &{72.09\small{}$\pm$0.7} &\textbf{62.54\small{}$\pm$0.7}  &90.90\small{}$\pm$0.6{\textcolor{red}{$^1$}}  &81.52\small{}$\pm$0.6 {\textcolor{red}{$^1$}} &50.79\small{}$\pm$0.6{\textcolor{red}{$^1$}} &\textbf{27.50\small{}$\pm$0.5}{\textcolor{red}{$^1$}} &\textbf{64.68} \\ 
NSAE(CE+CE)$^\ddagger$~\cite{liang2021boosting_cdfsl}
&68.51\small{}$\pm$0.8 &54.91\small{}$\pm$0.7 &71.02\small{}$\pm$0.7 &59.55\small{}$\pm$0.8  &{93.14\small{}$\pm$0.5} &{83.96\small{}$\pm$0.6} &\textbf{54.05\small{}$\pm$0.6} &{27.10\small{}$\pm$0.4} &{64.03} \\
RDC-FT{$^{*\ddagger}$} (ours)
&{67.77\small{}$\pm$0.4} &53.75\small{}$\pm$0.5 &\textbf{74.65\small{}$\pm$0.4} &60.63{\small{}$\pm$0.4}  &\textbf{93.55\small{}$\pm$0.3} &\textbf{84.67\small{}$\pm$0.3} &49.06\small{}$\pm$0.3 &25.48\small{}$\pm$0.2 &63.70 \\
\hline
\hline
\end{tabular}
}
\vspace{-0.1in}
\caption{\small{} {\textbf{Comparisons with SoTA methods}. The 5-way
    1/5-shot classification accuracies on 8 domains with
    \textbf{ResNet10} as the backbone. {$^\dagger$} indicates the
    result reported in~\cite{wang2021cross_ata}. {$^\ddagger$} \new{denotes}
    fine-tuning \new{in the} target domain. {*} \camera{\new{denotes
      using both} the labelled support data and the unlabelled query
      data}. $(\cdot)${\textcolor{red}{$^1$}} \new{denotes}
    our \new{reproduced} results \new{using} the official
    \new{released} code
    from~\cite{das2021distractor_cdfsl}. \textbf{Bold} \new{denotes the} best
    scores.}} 
\vspace{-0.1in}
\label{tab:sota}
\end{table*}

\section{Experiments}
\label{sec:exp}
\noindent\textbf{Dataset.}
Following the benchmarks in~\cite{wang2021cross_ata,liang2021boosting_cdfsl}, we used \textit{miniImageNet} as the source domain and another eight datasets
, \ie \textit{CUB}, \textit{Cars}, \textit{Places} \textit{Plantae}, \textit{CropDisease}, \textit{EuroSAT}, \textit{ISIC} and \textit{ChestX}, 
as target domains. In specific, \textit{miniImageNet}~\cite{vinyals2016matching} is a subset of of \textit{ILSVRC-2012}.
\textit{CUB}, \textit{Cars}, \textit{Places} and \textit{Plantae} are the target domains proposed in~\cite{tseng2020cross_fwt} for the evaluation on natural image domains, while \textit{CropDisease}, \textit{EuroSAT}, \textit{ISIC} and \textit{ChestX} are four domains proposed in~\cite{guo2020broader} for generalising the model to domains with different visual characteristics. For all experiments, we resized all the images to 224$\times$224 pixels and used data augmentations in~\cite{wang2021cross_ata,tseng2020cross_fwt} as image transformation.

\keypoint{Evaluation protocol.}
We followed the evaluation protocols in~\cite{wang2021cross_ata} to evaluate our method on CD-FSL. In specific, for each target domain, we randomly selected 2000 FSL tasks and each task contains 5 different classes. Each class has 1/5 support labelled data and additional 15 unlabelled data for evaluation the performance, formulating the 5-way 1/5-shot CD-FSL problem.
 {In all experiments, we reported the mean classification accuracy as well as 95\% confidence interval on the query set of each domain. For comprehensive comparison, we listed the average accuracy  {(shown as Ave. in Tab.~\ref{tab:baseline}~\ref{tab:sota} and~\ref{tab:dataaug})} of 8 domains.}

\keypoint{Implementation details.}
Following previous works~\cite{tseng2020cross_fwt,wang2021cross_ata,guo2020broader}, we used a ResNet10 as feature extractor.
Further, we used the same hyper-parameters for the experiments on different domains to fairly validate the generalisation ability. In specific, the feature extractor are pre-trained for 400 epochs on the base classes of \textit{mini}ImageNet with an Adam \camera{optimiser}. We set the learning rate as 0.001 and the batch size as 64. 
For our RDC method, we set $k=10$, $k_2=8$ and $\lambda=0.5$, and the reduced dimension $p$ was set as 64 for the non-linear subspace.
For the fine-tuning stage in RDC-FT, we set the attention scalar $\alpha=0.5$, temperature $\tau=3$ and $T=20$ epochs for model training using an Adam \camera{optimiser} with learning rate $\beta$ as 0.001.

\subsection{Comparison with baselines}
 {As our methods are based on a simple NPC classifier, here we start by comparing our RDC method with some baseline methods which also use a NPC classifier and do not need fine-tuning on a target domain. These baselines are: NPC that uses a NPC classifier on the pre-trained embedding, NPC+$l_2$ norm which utilises a NPC classifier on a $l_2$ normed feature embeddings, and ProtoNet~\cite{snell2017prototypical} that meta-learns a task-agnostic NPC classifier on \textit{mini}ImageNet. 
The results in Tab.~\ref{tab:baseline} show that RDC largely outperforms these baselines, boosting the simple NPC classifier to a strong one.} In particular, the performance on 1-shot learning is improved notably with $5\sim 10\%$ increases on the \textit{Ave.} accuracy compared to the baselines. This observation indicates that RDC is efficient to calibrate the distances by fully-leveraging the task information, \camera{\ie the labelled support data and unlabelled query data}.
 {We also note that the improvement on 5-shot is not as large as that on 1-shot. The reason is that the prototypes for the NPC classifier is more robust under many-shot setting, thus the original distances are less-biased and this calibration process improves less when the embedding is fixed. This limitation can be remitted by using the fine-tuning stage of our RDC-FT method.}

\subsection{Comparison with state-of-the-art methods}
\label{subsec:sota}
We further \camera{compared} our RDC-FT method with State-of-The-Art (SoTA) methods: 1) meta-learners:
GNN-FT~\cite{tseng2020cross_fwt} that meta-trains a GNN~\cite{satorras2018fewgnn} model with an additional Feature Transformation layer, GNN-LRP which uses a Layer-wise Relevance Propagation to guide the GNN model training, and TPN+ATA~\cite{wang2021cross_ata} that meta-learns TPN~\cite{liu2019fewTPN} with Adversarial Task Augmentation; 2) fine-tuning methods: a general Fine-tuning~\cite{guo2020broader} method, ConFT~\cite{das2021distractor_cdfsl} that fine-tunes model reusing the base classes, and NSAE~\cite{liang2021boosting_cdfsl} which pre-trains and fine-tunes model with an additional autoencoder task to improve the model generalisation.
From Tab.~\ref{subsec:sota},  {we observe that our RDC-FT method is superior to the SoTA methods on the 1-shot learning and competitive to SoTAs on the 5-shot learning.}  {Also, we notice that the performance is not superior to ConFT and NSAE methods for the 5-shot learning. The behind reasons are: 1) our method explores the task information in an unsupervised way while the others focus on fine-tuning with more labelled data; thus these methods benefit a lot from the 5-shot setting.  2) ConFT reuses more data from base classes for model fine-tuning. Thus the similar classes between source and target domain, \eg birds, cars, help to build more robust decision boundaries when model learning on related target domains, \eg \textit{CUB}, \textit{Cars}. But this approach requires more data and expensive computing resources. 3) NASE adopted an autoencoder to implicitly augment data to pre-train a generalisable model, and our method is theoretically orthogonal to this method for solving the CD-FSL problem.}

\subsection{Ablation study}
\noindent\textbf{Component analysis.}
{To investigate the efficacy of different components in RDC-FT, we \camera{ablated} the contribution of each element in RDC-FT: RDC \textit{w/o} subspace, RDC (in two spaces), RDC-FT \textit{w/o} subspace and RDC-FT (in two spaces). As in Tab.~\ref{tab:components}, a simple RDC process without subspace learning, which calibrates the distances only on the pre-trained representation, largely boosts the baseline NPC classifier by 7.99\% (1.55 \%) improvement on 1-shot (5-shot). The fine-tuning process, as in results of RDC-FT \textit{w/o} subspace, can 
\new{increase} the improvement \camera{by \new{an} iterative mapping
  process}, achieving 11.31\% (5.49 \%) improvement on 1-shot
(5-shot). 
Interestingly, we observe that the contribution of subspace for RDC ($\uparrow$0.94\% on 1-shot) is larger than that for RDC-FT ($\uparrow$0.50\% on 1-shot). This indicates that fine-tuning process can gradually alleviate the bias of pre-trained representations, thus the benefit of subspace becomes less in RDC-FT.} 
\begin{table}[t]
\centering
\scalebox{0.7}{
{
\begin{tabular}{lll}
\hline
Method & 5-way 1-shot & 5-way 5-shot \\
\hline
Baseline NPC                 &39.72     &57.42    \\
+RDC \textit{w/o} subspace   &47.71($\uparrow$7.99\%)     &58.97($\uparrow$1.55\%)  \\
+RDC    &48.65($\uparrow${8.93\%})      &60.06($\uparrow$2.64\%)      \\
+RDC-FT \textit{w/o} subspace         &51.03($\uparrow$11.31\%)      &62.91($\uparrow$5.49\%)      \\
+RDC-FT &51.53($\uparrow$11.81\%)      &63.70($\uparrow$6.28\%)       \\
\hline
\end{tabular}
}
}
\vspace{-0.1in}
\caption{\small \textbf{Component analysis of the proposed RDC-FT method.} The results are the average accuracies of 8 target domains. ($\uparrow \gamma$ \%) indicates $\gamma$\% improvement compared to the NPC baseline.}
\vspace{-0.15in}
    \label{tab:components}
\end{table}

\begin{table}[t]
\centering
\scalebox{0.7}{
{
\begin{tabular}{lcccccc}
\hline
\multirow{2}{*}{Method} &\multicolumn{6}{c}{{5-way 1-shot}} \\
\cline{2-7}
    &N/A   &linear &\small{}Gaussian & Poly. &\small{}Sigmoid &Ours \\
\hline
NPC+$l_2$ norm& 43.72 &45.54 &45.53  &45.49  & 45.59 &\textbf{45.81}   \\
RDC&47.77 &47.41 &47.38  &47.32  &47.46  &\textbf{48.36}    \\
\hline
\hline
\multirow{2}{*}{Method} &\multicolumn{6}{c}{{5-way 5-shot}} \\
\cline{2-7}
    &N/A   &linear &\small{}Gaussian & Poly. &\small{}Sigmoid &Ours \\
\hline
NPC+$l_2$ norm&59.03 &58.50 &58.52  &58.43  &58.56  &\textbf{59.72}    \\
RDC   &59.09 &58.83 &58.82  &58.75  &58.86  &\textbf{59.83}  \\
\hline
\end{tabular}
}
}
\vspace{-0.1in}
\caption{\small  {\textbf{Comparisons of different subspaces.} The average accuracies of 8 target domains by using NPC+$l_2$ norm and RDC methods on the subspaces constructed by KPCA with different kernels. N/A represents the original representation without subspace.}}
\vspace{-0.18in}
    \label{tab:kernel}
\end{table}

\keypoint{Comparison of different PCA methods.}
We compared our non-linear subspace to the subspaces constructed by Kernel PCA (KPCA) methods with different kernel types (linear, Gaussian, Polynomial and Sigmoid). For fair comparison, the dimensions of subspaces are set as 64 and we used the default parameters of KPCA methods in \textit{scikit-learn}~\cite{pedregosa2011scikit}.
Table~\ref{tab:kernel} shows that our non-linear subspace performs better than others. In particular, without RDC method, the KPCA methods can largely improve the performance (compared to N/A) on 1-shot learning, but the results of KPCA methods are just competitive to the original space when applying RDC on different subspaces. However, our non-linear subspace achieves consistent and stable improvement both with and \textit{w/o} RDC method, verifying the robustness and superiority of our non-linear subspace.

\begin{figure}[t]
    \centering
    \begin{subfigure}{.45\linewidth}
        \includegraphics[scale=0.31]{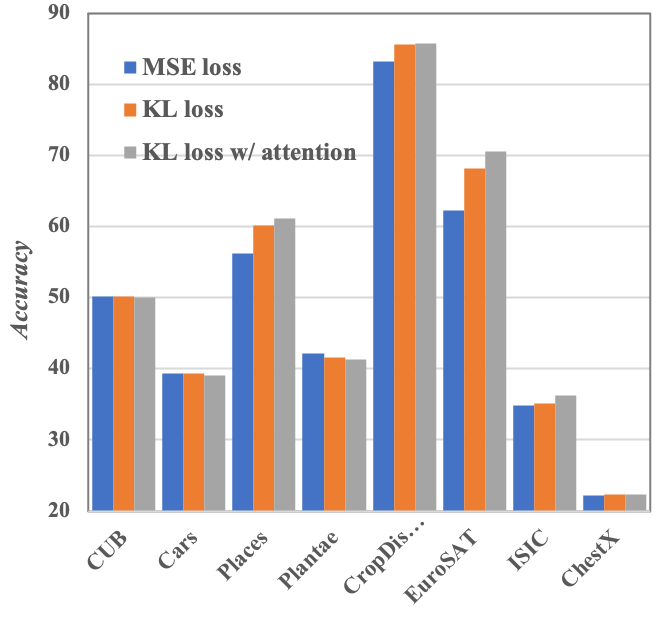}
        \vspace{-0.05in}
        \caption{5-way 1-shot evaluation.}
    \end{subfigure}
    \hskip1em
    \begin{subfigure}{.45\linewidth}
        \includegraphics[scale=0.3]{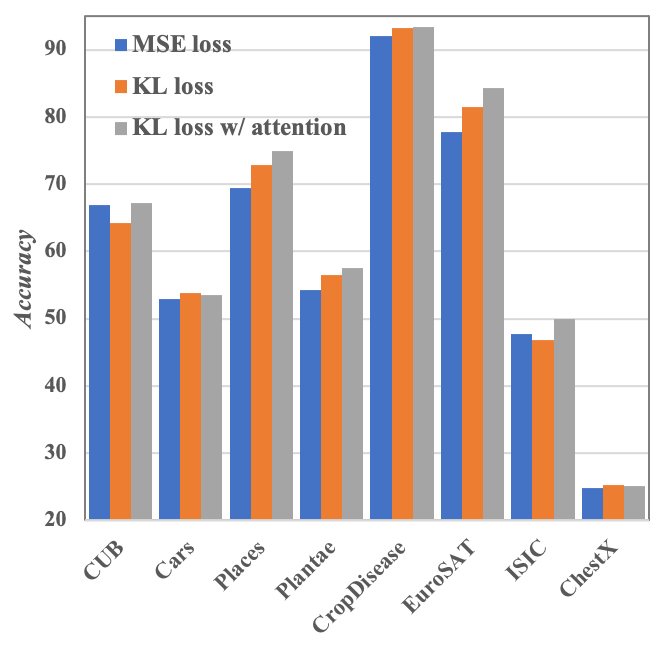}
        \vspace{-0.05in}
        \caption{5-way 5-shot evaluation.}
    \end{subfigure}
      \vspace{-0.1in}
    \caption{\small{} \textbf{Results of RDC-FT with different loss functions}. The evaluation is conducted on 2000 tasks from 8 target domains.}
    \vspace{-0.2in}
    \label{fig:loss}
\end{figure}

\keypoint{Effect of loss choices.}
 {
We evaluated the performance of RDC-FT with different loss functions. The results in Fig.~\ref{fig:loss} show that these three losses achieve competitive performance on \textit{CUB}, \textit{Cars}, \textit{ISIC} and \textit{chestX}, while the KL loss performs mostly better than MSE loss on \textit{Places}, \textit{Plantae}, \textit{EuroSAT}. These observations suggest the superiority of mapping the distance matrix in softened distributions. We conjecture this should attribute to the softening process which can alleviate the negative effect of the calibrated distances. Moreover, we note that the performance of KL loss can be further improved by an attention strategy on the distance matrices, verifying the efficacy of employing the expanded $k$-nearest neighbours list as an attention reference.
}

\keypoint{Visualisation.} 
\camera{To intuitively understand the advantages and limitations of
  the proposed RDC and RDC-FT methods, we used
  t-SNE~\cite{maaten2008visualizing} to display the classification
  results of \new{three} different methods \new{--} a NPC classifier, RDC,
  and RDC-FT, on a 5-way 1-shot FSL task from
  \textit{CropDisease}. The results in Fig.~\ref{fig:visual_rdc}
  \new{show}: (1) RDC can correct some misclassified samples that
  are \new{near} to the support exemplars, \ie the samples in red solid
  rectangles in plots(II)\&(III). However, RDC cannot \new{address effectively} the
  misclassified samples between different support exemplars, \ie the
  failure cases in plot(III). (2) From the samples in red dashed
  rectangles of plots(I)\&(IV), it is evident that RDC-FT can calibrate the
  distance-based distributions in the representational space,
  encouraging the feature representations to have \new{smaller} intra-class
  variations and \new{greater} inter-class margins, \new{resulting in fine-tuned
  representations being} more discriminative for classification. (3) The
  failure cases of RDC, \ie M-R1, M-R3, M-R4, and M-R5 in plot(III),
  can be correctly classified by RDC-FT with a simple NPC classifier,
  \new{as shown} in plot(V). This verifies the superiority of RDC-FT that gradually
  embeds the calibration information to the representational space. 
}

\begin{figure}[t]
  \centering
    \includegraphics[width=0.95\linewidth]{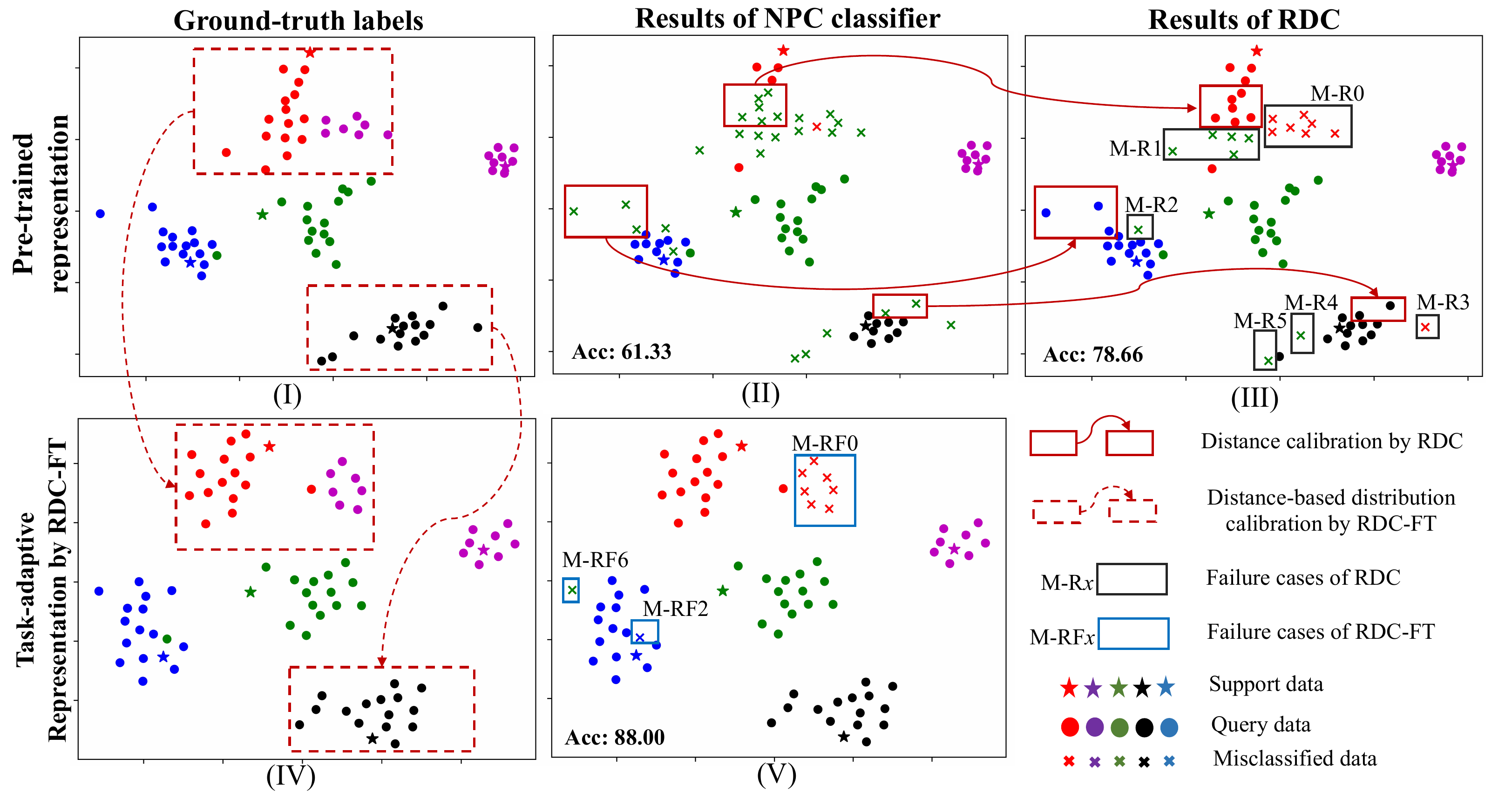}
    \vspace{-0.1in}
   \caption{\small{}\camera{\textbf{T-SNE visualisation of 5-way 1-shot task from \textit{CropDisease}.} We show the classification results with the NPC classifier and our RDC method in different representations, \ie Pre-trained representations and task-adaptive representation by RDC-FT. The different colours of the points (round/cross/star) represent the ground-truth labels or the labels assigned by NPC/RDC.}}
    \label{fig:visual_rdc} 
  \vspace{-0.1in}
\end{figure}

\keypoint{Incorporate with other method.}
{As RDC is a post-processing method, it can flexibly combine with other methods. Here we employed RDC on a general data augmentation method~\cite{yeh2020large}. The results in Tab.~\ref{tab:dataaug} indicate that RDC can achieve consistent improvement on other method, showing its generalisable ability. Currently we cannot evaluate our method on~\cite{liang2021boosting_cdfsl,liu2020feature} until their code is released.}


\begin{table}[t]
\centering
\scalebox{0.66}{
\begin{tabular}{lcccccccc}
\hline
\multirow{2}{*}{Method} &\multicolumn{8}{c}{{5-way 1-shot}} \\
\cline{2-9}
&\small{}CUB &\small{}Cars &\small{}Places &\small{}Plantae &\small{}Crop. &\small{}Euro. &\small{}ISIC &\small{}ChestX \\
\hline
NPC+$l_2$+DA$^{\dagger}$ 
&40.49 &34.53 &45.86 &36.78 &71.94 &64.29 &32.91 &\textbf{22.02} \\
RDC+DA$^{\dagger}$ 
&\textbf{45.98} &\textbf{38.17} &\textbf{58.30} &\textbf{40.73} &\textbf{84.73} &\textbf{71.59} &\textbf{34.44} &22.03  \\
\hline
\hline
\multirow{2}{*}{Method} &\multicolumn{8}{c}{{5-way 1-shot}} \\
\cline{2-9}
&\small{}CUB &\small{}Cars &\small{}Places &\small{}Plantae &\small{}Crop. &\small{}Euro. &\small{}ISIC &\small{}ChestX \\
\hline
NPC+$l_2$+DA$^{\dagger}$ 
&65.23 &53.61 &68.91 &57.72 &91.66 &83.09 &50.63 &\textbf{26.03} \\ 
RDC+DA$^{\dagger}$ 
&\textbf{67.68} &\textbf{55.13} &\textbf{73.06} &\textbf{60.34} &\textbf{93.41} &\textbf{84.67} &\textbf{51.53} &{25.79} \\
\hline
\end{tabular}
}
\vspace{-0.1in}
\caption{\small{}\textbf{Incorporate RDC with others.} $^{\dagger}$ denotes using NPC \emph{w/} $l_{2}$ norm on the model fine-tuned with data augmentation in~\cite{yeh2020large}.}
\vspace{-0.15in}
\label{tab:dataaug}
\end{table}

\section{Conclusions}
In this paper, we proposed a Ranking Distance Calibration (RDC) method to calibrate the biased distances in CD-FSL. 
The calibration process is achieved by a re-ranking method with a $k$-reciprocal discovery and encoding process. 
As the pre-trained linear embedding is biased for target domain, we further proposed a non-linear subspace followed by a calibration process on it. 
Our RDC method averages the calibrated distances on the two spaces to a robust distance matrix.
Moreover, we introduced a RDC-FT method to fine-tune the embedding with the calibrated distances, yielding a discriminative representation for CD-FSL task.

\keypoint{Limitation.} As the image retrieval perspective of our approach is to discover the task information unsupervised, exploring comprehensive leveraging of the label information and the task information should be considered, especially in the many-shot cases, \eg 5-shot. 

\keypoint{Ackowledgement.}
\camera{
This work was supported by Vision Semantics Limited, the Alan Turing
Institute, the China Scholarship Council, and the NSFC under Grant
(No. 62076067). We shall also \new{thank Jun} Liu and Bin-Bin Gao of
Tencent YouTu Lab for helpful discussions and GPU support. 
}

{\small
\bibliographystyle{ieee_fullname}
\bibliography{egbib}

\begin{thebibliography}{10}\itemsep=-1pt

\bibitem{adler2020cross}
Thomas Adler, Johannes Brandstetter, Michael Widrich, Andreas Mayr, David
  Kreil, Michael Kopp, G{\"u}nter Klambauer, and Sepp Hochreiter.
\newblock Cross-domain few-shot learning by representation fusion.
\newblock {\em arXiv preprint arXiv:2010.06498}, 2020.

\bibitem{bai2016sparse}
Song Bai and Xiang Bai.
\newblock Sparse contextual activation for efficient visual re-ranking.
\newblock {\em IEEE TIP}, 25(3):1056--1069, 2016.

\bibitem{chen2019closer}
Wei-Yu Chen, Yen-Cheng Liu, Zsolt Kira, Yu-Chiang~Frank Wang, and Jia-Bin
  Huang.
\newblock A closer look at few-shot classification.
\newblock In {\em ICLR}, 2019.

\bibitem{chum2007total}
Ondrej Chum, James Philbin, Josef Sivic, Michael Isard, and Andrew Zisserman.
\newblock Total recall: Automatic query expansion with a generative feature
  model for object retrieval.
\newblock In {\em ICCV}, pages 1--8. IEEE, 2007.

\bibitem{das2022confess}
Debasmit Das, Sungrack Yun, and Fatih Porikli.
\newblock Confe{SS}: A framework for single source cross-domain few-shot
  learning.
\newblock In {\em ICLR}, 2022.

\bibitem{das2021distractor_cdfsl}
Rajshekhar Das, Yu-Xiong Wang, and Jose~MF Moura.
\newblock On the importance of distractors for few-shot classification.
\newblock In {\em ICCV}, pages 9030--9040, 2021.

\bibitem{fang2021kernel_hyperbolic}
Pengfei Fang, Mehrtash Harandi, and Lars Petersson.
\newblock Kernel methods in hyperbolic spaces.
\newblock In {\em ICCV}, pages 10665--10674, 2021.

\bibitem{finn2017model}
Chelsea Finn, Pieter Abbeel, and Sergey Levine.
\newblock Model-agnostic meta-learning for fast adaptation of deep networks.
\newblock In {\em ICML}, pages 1126--1135, 2017.

\bibitem{fu2021metamixup}
Yuqian Fu, Yanwei Fu, and Yu-Gang Jiang.
\newblock Meta-fdmixup: Cross-domain few-shot learning guided by labeled target
  data.
\newblock {\em ACM MM}, pages 5326--5334, 2021.

\bibitem{guo2020broader}
Yunhui Guo, Noel~C Codella, Leonid Karlinsky, James~V Codella, John~R Smith,
  Kate Saenko, Tajana Rosing, and Rogerio Feris.
\newblock A broader study of cross-domain few-shot learning.
\newblock In {\em ECCV}, pages 124--141. Springer, 2020.

\bibitem{he2004manifold_rank}
Jingrui He, Mingjing Li, Hong-Jiang Zhang, Hanghang Tong, and Changshui Zhang.
\newblock Manifold-ranking based image retrieval.
\newblock In {\em ACM MM}, pages 9--16, 2004.

\bibitem{hu2022switch}
Zhengdong Hu, Yifan Sun, and Yi Yang.
\newblock Switch to generalize: Domain-switch learning for cross-domain
  few-shot classification.
\newblock In {\em International Conference on Learning Representations}, 2022.

\bibitem{huang2015cross_retrivel}
Junshi Huang, Rogerio~S Feris, Qiang Chen, and Shuicheng Yan.
\newblock Cross-domain image retrieval with a dual attribute-aware ranking
  network.
\newblock In {\em ICCV}, pages 1062--1070, 2015.

\bibitem{khrulkov2020hyperbolic}
Valentin Khrulkov, Leyla Mirvakhabova, Evgeniya Ustinova, Ivan Oseledets, and
  Victor Lempitsky.
\newblock Hyperbolic image embeddings.
\newblock In {\em CVPR}, pages 6418--6428, 2020.

\bibitem{kim2021comparing}
Taehyeon Kim, Jaehoon Oh, NakYil Kim, Sangwook Cho, and Se-Young Yun.
\newblock Comparing kullback-leibler divergence and mean squared error loss in
  knowledge distillation.
\newblock {\em IJCAI}, 2021.

\bibitem{li2020few_self}
Jianyi Li and Guizhong Liu.
\newblock Few-shot image classification via contrastive self-supervised
  learning.
\newblock {\em arXiv preprint arXiv:2008.09942}, 2020.

\bibitem{pan2021mfl}
Pan Li, Yanwei Fu, and Shaogang Gong.
\newblock Regularising knowledge transfer by meta functional learning.
\newblock In {\em IJCAI}, 2021.

\bibitem{li2021simple}
Pan Li, Da Li, Wei Li, Shaogang Gong, Yanwei Fu, and Timothy~M Hospedales.
\newblock A simple feature augmentation for domain generalization.
\newblock In {\em ICCV}, pages 8886--8895, 2021.

\bibitem{li2021plain}
Pan Li, Guile Wu, Shaogang Gong, and Xu Lan.
\newblock Semi-supervised few-shot learning with pseudo label refinement.
\newblock In {\em ICME}, pages 1--6. IEEE, 2021.

\bibitem{liang2021boosting_cdfsl}
Hanwen Liang, Qiong Zhang, Peng Dai, and Juwei Lu.
\newblock Boosting the generalization capability in cross-domain few-shot
  learning via noise-enhanced supervised autoencoder.
\newblock In {\em ICCV}, pages 9424--9434, October 2021.

\bibitem{liu2020feature}
Bingyu Liu, Zhen Zhao, Zhenpeng Li, Jianan Jiang, Yuhong Guo, and Jieping Ye.
\newblock Feature transformation ensemble model with batch spectral
  regularization for cross-domain few-shot classification.
\newblock {\em arXiv preprint arXiv:2005.08463}, 2020.

\bibitem{liu2013pop_rank}
Chunxiao Liu, Chen~Change Loy, Shaogang Gong, and Guijin Wang.
\newblock Pop: Person re-identification post-rank optimisation.
\newblock In {\em CVPR}, pages 441--448, 2013.

\bibitem{liu2020urt}
Lu Liu, William~L Hamilton, Guodong Long, Jing Jiang, and Hugo Larochelle.
\newblock A universal representation transformer layer for few-shot image
  classification.
\newblock In {\em ICLR}, 2021.

\bibitem{liu2019fewTPN}
Yanbin Liu, Juho Lee, Minseop Park, Saehoon Kim, Eunho Yang, Sungju Hwang, and
  Yi Yang.
\newblock Learning to propagate labels: Transductive propagation network for
  few-shot learning.
\newblock In {\em ICLR}, 2019.

\bibitem{liu2018adaptivererank}
Yong Liu, Lin Shang, and Andy Song.
\newblock Adaptive re-ranking of deep feature for person re-identification.
\newblock {\em arXiv preprint arXiv:1811.08561}, 2018.

\bibitem{loy2013person_rank}
Chen~Change Loy, Chunxiao Liu, and Shaogang Gong.
\newblock Person re-identification by manifold ranking.
\newblock In {\em ICIP}, pages 3567--3571. IEEE, 2013.

\bibitem{maaten2008visualizing}
Laurens van~der Maaten and Geoffrey Hinton.
\newblock Visualizing data using t-sne.
\newblock 9(Nov):2579--2605, 2008.

\bibitem{oh2021boil}
Jaehoon Oh, Hyungjun Yoo, ChangHwan Kim, and Se-Young Yun.
\newblock Boil: Towards representation change for few-shot learning.
\newblock In {\em ICLR}, 2021.

\bibitem{pedregosa2011scikit}
Fabian Pedregosa, Ga{\"e}l Varoquaux, Alexandre Gramfort, Vincent Michel,
  Bertrand Thirion, Olivier Grisel, Mathieu Blondel, Peter Prettenhofer, Ron
  Weiss, Vincent Dubourg, et~al.
\newblock Scikit-learn: Machine learning in python.
\newblock {\em the Journal of machine Learning research}, 12:2825--2830, 2011.

\bibitem{phoo2021STARTUP}
Cheng~Perng Phoo and Bharath Hariharan.
\newblock Self-training for few-shot transfer across extreme task differences.
\newblock In {\em ICLR}, 2021.

\bibitem{qin2011hello_kreciprocal}
Danfeng Qin, Stephan Gammeter, Lukas Bossard, Till Quack, and Luc Van~Gool.
\newblock Hello neighbor: Accurate object retrieval with k-reciprocal nearest
  neighbors.
\newblock In {\em CVPR}, pages 777--784. IEEE, 2011.

\bibitem{ravi2016optimization}
Sachin Ravi and Hugo Larochelle.
\newblock Optimization as a model for few-shot learning.
\newblock In {\em ICLR}, 2017.

\bibitem{sarfraz2018posererank}
M~Saquib Sarfraz, Arne Schumann, Andreas Eberle, and Rainer Stiefelhagen.
\newblock A pose-sensitive embedding for person re-identification with expanded
  cross neighborhood re-ranking.
\newblock In {\em CVPR}, pages 420--429, 2018.

\bibitem{satorras2018fewgnn}
Victor~Garcia Satorras and Joan~Bruna Estrach.
\newblock Few-shot learning with graph neural networks.
\newblock In {\em ICLR}, 2019.

\bibitem{simon2020adaptive}
Christian Simon, Piotr Koniusz, Richard Nock, and Mehrtash Harandi.
\newblock Adaptive subspaces for few-shot learning.
\newblock In {\em CVPR}, pages 4136--4145, 2020.

\bibitem{snell2017prototypical}
Jake Snell, Kevin Swersky, and Richard Zemel.
\newblock Prototypical networks for few-shot learning.
\newblock In {\em NeurIPS}, pages 4077--4087, 2017.

\bibitem{sun2021explanationcdfsl}
Jiamei Sun, Sebastian Lapuschkin, Wojciech Samek, Yunqing Zhao, Ngai-Man
  Cheung, and Alexander Binder.
\newblock Explanation-guided training for cross-domain few-shot classification.
\newblock In {\em ICIP}, pages 7609--7616. IEEE, 2021.

\bibitem{sung2018learning}
Flood Sung, Yongxin Yang, Li Zhang, Tao Xiang, Philip~HS Torr, and Timothy~M
  Hospedales.
\newblock Learning to compare: Relation network for few-shot learning.
\newblock In {\em CVPR}, pages 1199--1208, 2018.

\bibitem{tian2020rethinking}
Yonglong Tian, Yue Wang, Dilip Krishnan, Joshua~B Tenenbaum, and Phillip Isola.
\newblock Rethinking few-shot image classification: a good embedding is all you
  need?
\newblock In {\em ECCV}, pages 266--282. Springer, 2020.

\bibitem{triantafillou2017fewretrivel}
Eleni Triantafillou, Richard Zemel, and Raquel Urtasun.
\newblock Few-shot learning through an information retrieval lens.
\newblock In {\em NeurIPS}, pages 2252--2262, 2017.

\bibitem{tseng2020cross_fwt}
Hung-Yu Tseng, Hsin-Ying Lee, Jia-Bin Huang, and Ming-Hsuan Yang.
\newblock Cross-domain few-shot classification via learned feature-wise
  transformation.
\newblock In {\em ICLR}, 2020.

\bibitem{van2008visualizing}
Laurens Van~der Maaten and Geoffrey Hinton.
\newblock Visualizing data using t-sne.
\newblock {\em Journal of machine learning research}, 9(11), 2008.

\bibitem{vinyals2016matching}
Oriol Vinyals, Charles Blundell, Timothy Lillicrap, Daan Wierstra, et~al.
\newblock Matching networks for one shot learning.
\newblock {\em NeurIPS}, 29:3630--3638, 2016.

\bibitem{volpi2018generalizing_dataaug}
Riccardo Volpi, Hongseok Namkoong, Ozan Sener, John Duchi, Vittorio Murino, and
  Silvio Savarese.
\newblock Generalizing to unseen domains via adversarial data augmentation.
\newblock {\em NeurIPS}, 2018.

\bibitem{wang2020high}
Guan'an Wang, Shuo Yang, Huanyu Liu, Zhicheng Wang, Yang Yang, Shuliang Wang,
  Gang Yu, Erjin Zhou, and Jian Sun.
\newblock High-order information matters: Learning relation and topology for
  occluded person re-identification.
\newblock In {\em CVPR}, pages 6449--6458, 2020.

\bibitem{wang2021cross_ata}
Haoqing Wang and Zhi-Hong Deng.
\newblock Cross-domain few-shot classification via adversarial task
  augmentation.
\newblock {\em IJCAI}, 2021.

\bibitem{Wu_2021_ICCV}
Guile Wu and Shaogang Gong.
\newblock Collaborative optimization and aggregation for decentralized domain
  generalization and adaptation.
\newblock In {\em ICCV}, pages 6484--6493, October 2021.

\bibitem{xi2021_reranking}
Shell Xu Hu Othman Sbai Mathieu~Aubry Xi~Shen, Yang~Xiao.
\newblock Re-ranking for image retrieval and transductive few-shot
  classification.
\newblock {\em NeurIPS}, 2021.

\bibitem{yan2021unsupervised_hyperbolic}
Jiexi Yan, Lei Luo, Cheng Deng, and Heng Huang.
\newblock Unsupervised hyperbolic metric learning.
\newblock In {\em CVPR}, pages 12465--12474, 2021.

\bibitem{yeh2020large}
Jia-Fong Yeh, Hsin-Ying Lee, Bing-Chen Tsai, Yi-Rong Chen, Ping-Chia Huang, and
  Winston~H Hsu.
\newblock Large margin mechanism and pseudo query set on cross-domain few-shot
  learning.
\newblock {\em arXiv preprint arXiv:2005.09218}, 2020.

\bibitem{yoon2019tapnet}
Sung~Whan Yoon, Jun Seo, and Jaekyun Moon.
\newblock Tapnet: Neural network augmented with task-adaptive projection for
  few-shot learning.
\newblock In {\em ICML}, pages 7115--7123. PMLR, 2019.

\bibitem{zhang2020deepemd}
Chi Zhang, Yujun Cai, Guosheng Lin, and Chunhua Shen.
\newblock Deepemd: Few-shot image classification with differentiable earth
  mover's distance and structured classifiers.
\newblock In {\em CVPR}, pages 12203--12213, 2020.

\bibitem{NEURIPS2020_adver_dataaug}
Long Zhao, Ting Liu, Xi Peng, and Dimitris Metaxas.
\newblock Maximum-entropy adversarial data augmentation for improved
  generalization and robustness.
\newblock In {\em NeurIPS}, volume~33, pages 14435--14447, 2020.

\bibitem{zhong2017rerank}
Zhun Zhong, Liang Zheng, Donglin Cao, and Shaozi Li.
\newblock Re-ranking person re-identification with k-reciprocal encoding.
\newblock In {\em CVPR}, pages 1318--1327, 2017.

\bibitem{zhou2021dg_mixstyle}
Kaiyang Zhou, Yongxin Yang, Yu Qiao, and Tao Xiang.
\newblock Domain generalization with mixstyle.
\newblock In {\em ICLR}, 2021.

\end{thebibliography}
}

\clearpage


\begin{appendices}
\renewcommand\thesection{\Alph{section}}
\setcounter{figure}{0} 
\renewcommand\thefigure{\Alph{section}\arabic{figure}}   
\setcounter{table}{0}
\renewcommand\thetable{\alph{table}} 

\noindent\textbf{\Large{}Supplementary Material}
\vspace{0.1in}

In this supplementary material, we present:
\begin{itemize}
    \item To validate the robustness of the proposed RDC and RDC-FT methods, we analyse the sensitivities of different hyper-parameters, \ie $\lambda$, $p$ and $\alpha$ (in Sec.~\ref{sec:hyper});
    \item To \camera{qualitatively show the effectiveness of the proposed RDC and RDC-FT methods, we show the ranking lists of a case study with and \emph{w/o} RDC~(Figure~\ref{fig:visual_rdc} in Sec.~\ref{sec:visual}), and further use t-SNE~\cite{maaten2008visualizing} to visulise the embeddings of three target domains with and \textit{w/o} RDC-FT (Figure~\ref{fig:visualisation} in Sec.~\ref{sec:visual}); }
    \item For better understanding the computing process of the Jaccard distance,  we illustrate the algorithm of the Jaccard distance in Sec.~\ref{sec:algorithm}.
    \item The notations for all symbols and hyper-parameters used in the main paper are defined (in Sec.~\ref{sec:symbol});
\end{itemize}

\section{Sensitivity analysis of the hyper-parameters}
\label{sec:hyper}
\begin{figure*}[t]
    \centering
    \vspace{-0.1in}
    \begin{subfigure}{.3\linewidth}
        \centering
        \includegraphics[scale=0.3]{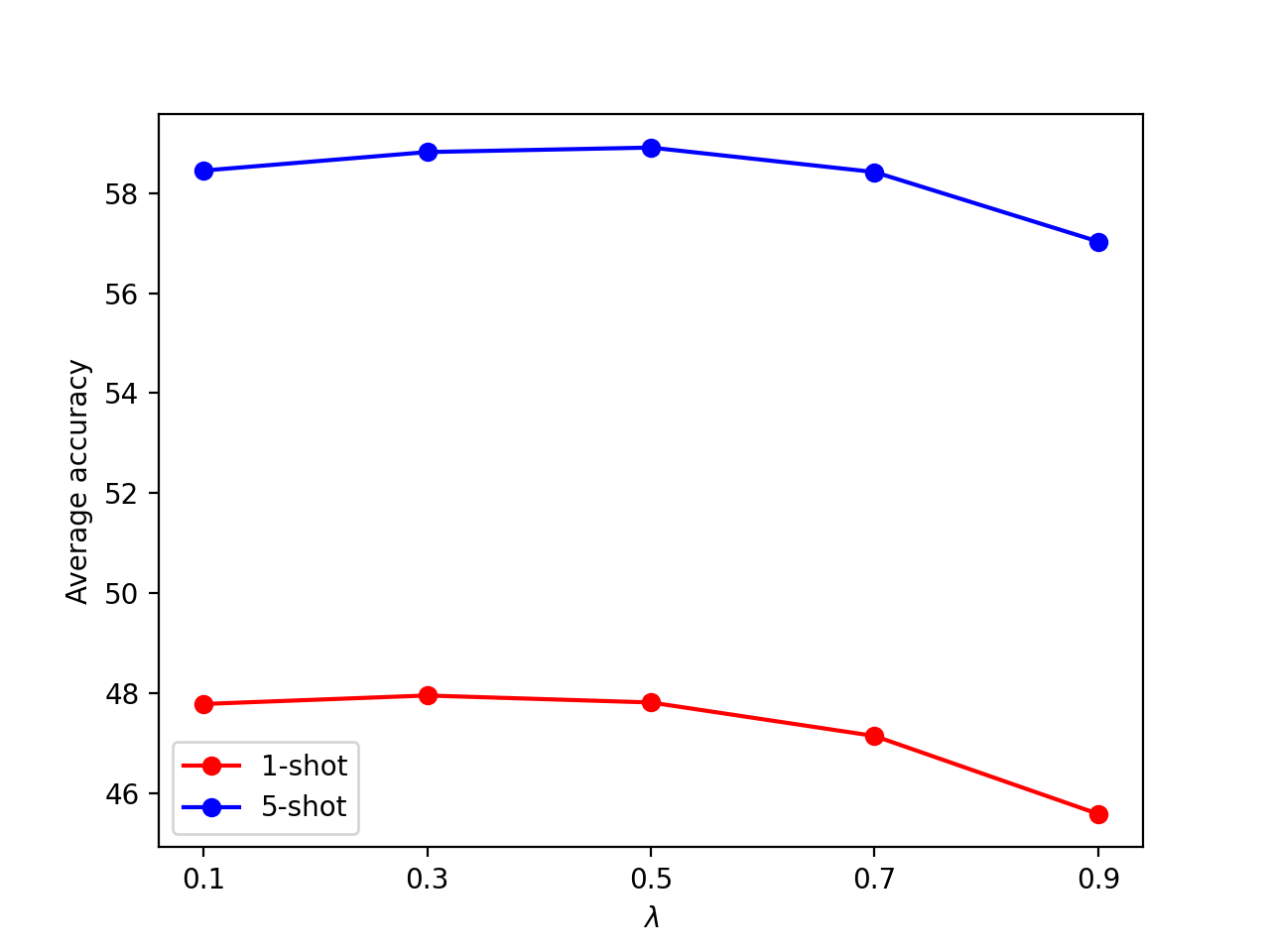}
        \vspace{-0.01in}
        \caption{Sensitivity analysis of $\lambda$.}
    \end{subfigure}
    \begin{subfigure}{.3\linewidth}
        \centering
        \includegraphics[scale=0.3]{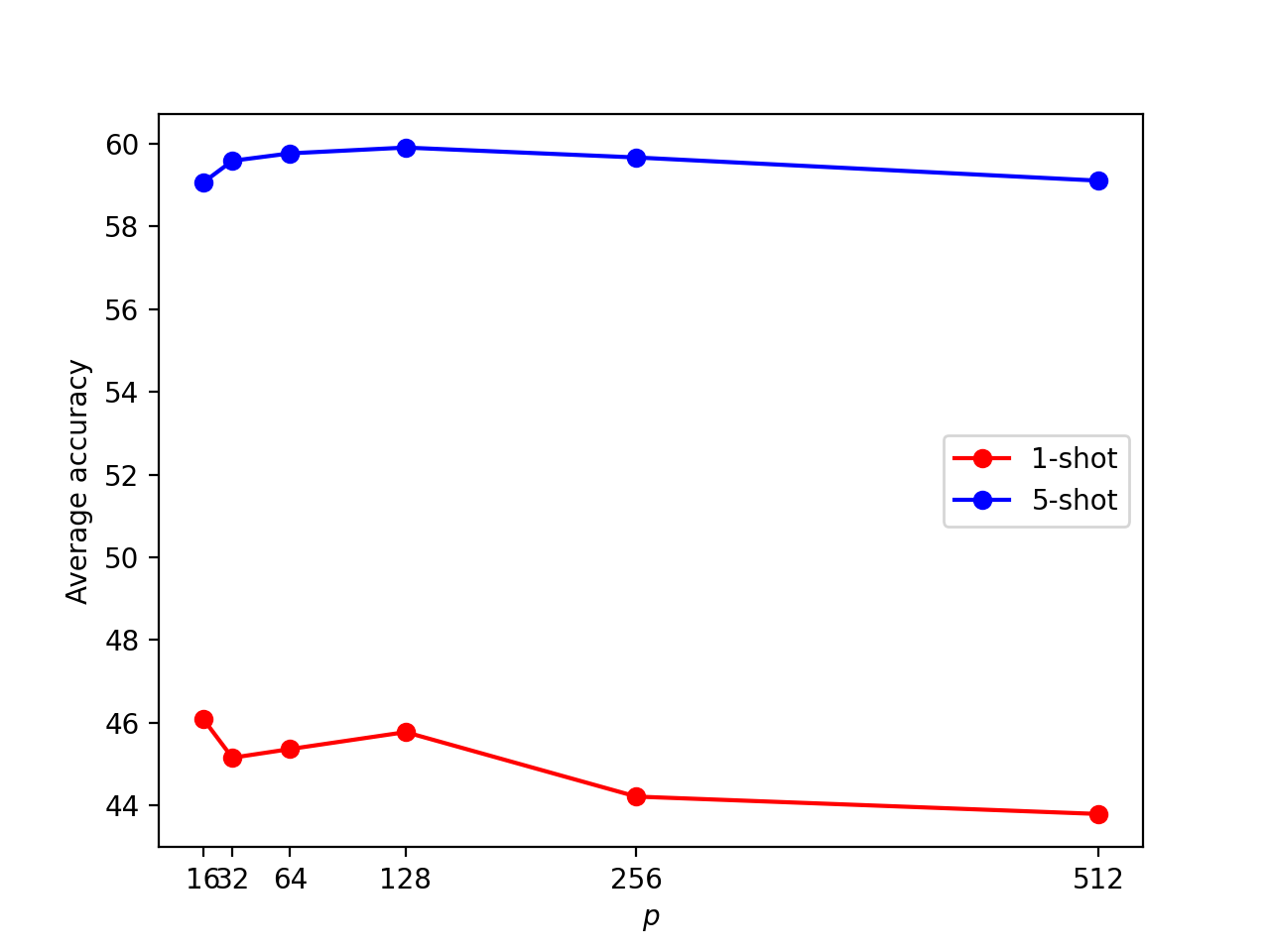}
        \vspace{-0.01in}
        \caption{Sensitivity analysis of $p$.}
    \end{subfigure}
    \begin{subfigure}{.3\linewidth}
        \centering
        \includegraphics[scale=0.28]{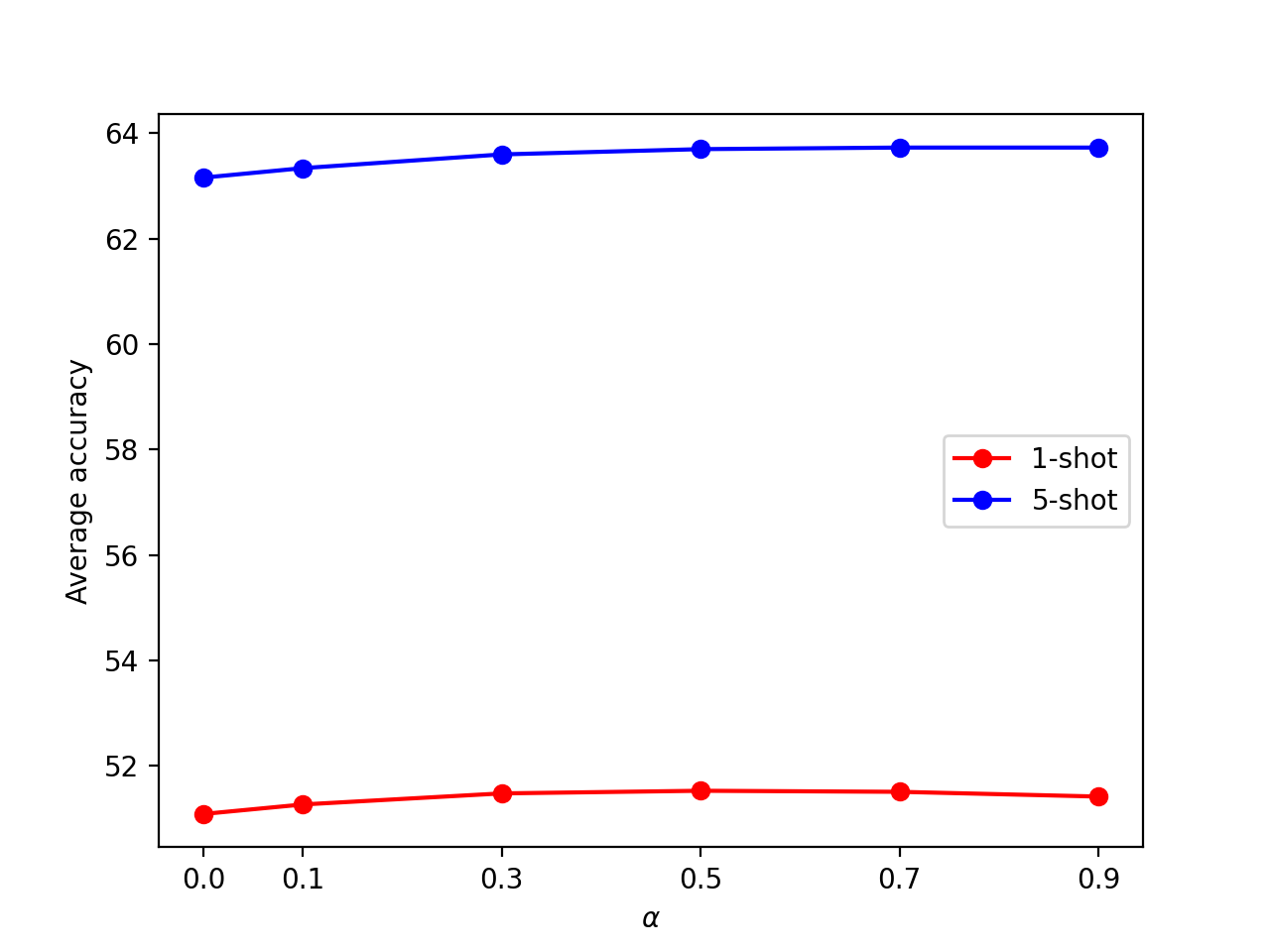}
        \vspace{-0.01in}
        \caption{Sensitivity analysis of $\alpha$.}
    \end{subfigure}
      \vspace{-0.1in}
    \caption{\small{} \textbf{Analysis of the hyper-parameters, \ie $\lambda$, $p$ and $\alpha$.} The evaluation results are the average accuracies of 8 target domains.}
    \vspace{-0.07in}
    \label{fig:parameters}
\end{figure*}

In all experiments of the main paper, we reported the results on 8 target domains with the same hyper-parameters. In practice, our method is robust to the hyper-parameters selection as shown in Fig.~\ref{fig:parameters}. Further, we analyse in depth three key hyper-parameters, $\lambda$, $p$  and $\alpha$.

\subsection{Effect of the trade-off scalar $\lambda$}
The trade-off scalar $\lambda$ is used to balance the original distance and the Jaccard distance for the proposed RDC method, thus it is a critical hyper-parameter for RDC.
We conducted experiments to test RDC with $\lambda=\{0.1,0.3,0.5,0.7,0.9\}$ on the pre-trained space. 

The results are shown in Tab.~\ref{tab:lambda} and Fig.~\ref{fig:parameters}(a), from which we can see that assigning smaller weights to the original distance (smaller $\lambda$) is a better choice for RDC. In particular, the best $\lambda$ for 1-shot is 0.3 while that for 5-shot is 0.5. This indicates that the original distance becomes more robust when the shot increases, thus the original space should occupy larger weights in the calibrated distance. 
Besides, when $\lambda$ is between 0.1 to 0.5, the average accuracies of RDC are stable, verifying the robustness of $\lambda$. Therefore, \camera{we set $\lambda=0.5$ in all experiments of the main paper.}

\begin{table}[]
\centering
\scalebox{0.63}{
\begin{tabular}{cccccccccc}
\toprule
\multicolumn{1}{c}{\multirow{2}{*}{$\lambda$}} &
\multicolumn{9}{c}{5-way 1-shot}                                            \\
\cline{2-10}
\multicolumn{1}{c}{}                        
&CUB  &Car  &Places &Plantae &Crop  &Euro. &ISIC  &Chestx & Ave.  \\
\midrule
0.1&47.10     & 37.01       &57.52  & 41.06   &\textbf{80.15} &\textbf{65.66}   & 31.34 & 22.40  &47.78 \\
0.3&{47.46}   & 37.47       &\textbf{57.67}  &\textbf{41.29}   & 79.82 &\textbf{65.67}   & 31.72 & 22.52  &\textbf{47.95} \\
0.5&\textbf{47.51} &{37.79} & 57.28  &\textbf{41.27}   & 78.87 & 65.21   & 31.92 & 22.61  &47.81 \\
0.7&47.12 &\textbf{37.85}   & 55.74  &40.92   & 76.71 & 64.13   &\textbf{31.98} & 22.70  &47.14 \\
0.9&45.94 & 37.37           &52.51  &40.12   & 71.72 & 62.19   &\textbf{31.98} &\textbf{22.78}  &45.58 \\
\midrule
\midrule
\multirow{2}{*}{$\lambda$}  & \multicolumn{9}{c}{5-way 5-shot}                                            \\
\cline{2-10}
&CUB  &Car  &Places &Plantae &Crop  &Euro. &ISIC  &Chestx & Ave.  \\
\midrule
0.1 &60.94 &49.42 &70.66  &55.79 &\textbf{88.67} &\textbf{76.95} &40.11 &25.03  &58.45 \\
0.3 &61.64 &50.25 &\textbf{71.00}  &\textbf{56.12} &{88.52} &\textbf{76.96} &40.72 &25.32  &58.82 \\
0.5 &\textbf{62.04} &50.90 &\textbf{71.01}  &56.10 &88.00 &76.54 &\textbf{41.00} &{25.66}  &\textbf{58.91} \\
0.7 &61.81 &\textbf{51.07} &70.30  &55.41 &86.74 &75.44 &40.86 &\textbf{25.75}  &58.42 \\
0.9 &60.60 &50.63 &67.97  &53.85 &83.70 &73.30 &40.45 &25.70  &57.03 \\
\bottomrule
\end{tabular}
}
\vspace{-0.1in}
\caption{\small{}\textbf{Analysis of the trade-off scalar $\lambda$.} Results of RDC with $\lambda=\{0.1,0.3,0.5,0.7,0.9\}$ on the pre-trained space.}
\label{tab:lambda}
\vspace{-0.1in}
\end{table}

\begin{table}[]
\centering
\scalebox{0.63}{
\begin{tabular}{cccccccccc}
\toprule
\multicolumn{1}{c}{\multirow{2}{*}{\textbf{$p$}}} &
\multicolumn{9}{c}{5-way 1-shot}                                            \\
\cline{2-10}
\multicolumn{1}{c}{}                        
&CUB  &Car  &Places &Plantae &Crop  &Euro. &ISIC  &Chestx & Ave.  \\
\midrule
16  &\textbf{48.56} &37.12 &\textbf{53.27} &\textbf{41.71} &\textbf{71.98} &{61.57} &\textbf{31.63} &22.81 &\textbf{46.08} \\
32  &47.43 &37.48 &51.38 &40.06 &69.93 &60.54 &31.45 &22.87 &45.14 \\
64  &47.65 &37.98 &52.07 &40.51 &69.95 &60.42 &31.45 &22.87 &45.36 \\
128 &47.78 &\textbf{38.27} &53.10 &{40.91} &70.37 &61.22 &31.60 &\textbf{22.90} &{45.77} \\
256 &44.88 &36.39 &48.54 &40.08 &67.56 &\textbf{61.70} &\textbf{31.63} &22.89 &44.21 \\
512 &44.33 &36.14 &48.75 &39.24 &66.76 &61.00 &31.34 &22.79 &43.79 \\
\midrule
\midrule
\multirow{2}{*}{\textbf{$p$}}  & \multicolumn{9}{c}{5-way 5-shot}                                            \\
\cline{2-10}
&CUB  &Car  &Places &Plantae &Crop  &Euro. &ISIC  &Chestx & Ave.  \\
\midrule
16  &65.00 &49.58 &70.46 &56.29 &\textbf{88.54} &76.50 &40.15 &26.01 &59.07 \\
32  &\textbf{65.38} &51.80 &70.29 &56.48 &88.41  &76.89 &41.12 &26.38 &59.59 \\
64  &65.19 &52.43 &70.55 &56.46 &88.27  &77.20 &41.74 &{26.46} &59.77 \\
128 &{64.93} &\textbf{52.77} &\textbf{71.21} &\textbf{56.59} & {88.27} &{77.20} &\textbf{41.81} &\textbf{26.49} & \textbf{59.91} \\
256 &63.94 &52.34 &69.99 &\textbf{56.58} &{88.41} &\textbf{77.81} &\textbf{41.81} &\textbf{26.49}&59.67 \\
512 &63.10 &52.13 &69.95 &55.39 &87.78 &76.88 &41.29 &26.35 &59.11\\
\bottomrule
\end{tabular}
}
\vspace{-0.1in}
\caption{\small{}\textbf{Analysis of the number of the reduced dimensions $p$.} Results of NPC on the proposed non-linear subspaces with $p=\{16,32,64,128,256,512\}$. }
\label{tab:p}
\vspace{-0.1in}
\end{table}

\subsection{Influence of the reduced dimensions $p$}
The dimensions $p$ in the subspace is a key parameter to build our non-linear space. Typically, we choose $p=\{16,32,64,128,256,512\}$ ($p=512$ represents the original space) to test the effects of different dimension $p$. 

Table~\ref{tab:p} and Fig.~\ref{fig:parameters}(b) show that the performance on different subspaces are stable when $p$ is smaller than 128. This observation shows that the subspaces constructed by the hyperbolic tangent transformation are not sensitive to the reduced dimensions.  In particular, the subspace with $p=16$ is the best dimension for 1-shot learning and that with $p=128$ is the best dimension for 5-shot learning. To make a balance among different shot learning, we set $p=64$ in all experiments of the main paper.

\begin{table}[]
\centering
\scalebox{0.65}{
\begin{tabular}{cccccccccc}
\toprule
\multicolumn{1}{c}{\multirow{2}{*}{$\alpha$}} &
\multicolumn{9}{c}{5-way 1-shot}                                            \\
\cline{2-10}
\multicolumn{1}{c}{}                        
&CUB  &Car  &Places &Plantae &Crop  &Euro. &ISIC  &Chestx & Ave.  \\
\midrule
0   &51.13 &39.15 &60.18 &43.93 &86.01 &70.71 &35.34 &22.29 &51.09 \\
0.1 &51.14 &\textbf{39.21} &60.63 &44.07 &86.16 &71.11 &35.56 &22.29 &51.27 \\
0.3 &\textbf{51.22} &39.17 &61.25 &\textbf{44.33} &86.31 &71.49 &35.79 &22.28 &51.48 \\
0.5 &51.20 &39.20 &\textbf{61.50} &\textbf{44.33} &\textbf{86.33} &\textbf{71.57} &35.84 &22.27 &\textbf{51.53} \\
0.7 &51.17 &39.17 &\textbf{61.50} &44.23 &86.24 &71.52 &35.98 &22.23 &51.51 \\
0.9 &50.97 &39.13 &61.41 &44.05 &86.05 &71.46 &\textbf{36.04} &22.23 &51.42 \\
\midrule
\midrule
\multirow{2}{*}{$\alpha$}  & \multicolumn{9}{c}{5-way 5-shot}                                            \\
\cline{2-10}
&CUB  &Car  &Places &Plantae &Crop  &Euro. &ISIC  &Chestx & Ave.  \\
\midrule
0   &67.35 &53.80 &73.62 &60.19 &93.25 &83.85 &47.63 &\textbf{25.58} &63.16 \\
0.1 &67.46 &53.81 &73.96 &60.48 &93.34 &84.14 &48.02 &25.54 &63.34 \\
0.3 &67.65 &53.80 &74.42 &\textbf{60.72} &93.48 &84.51 &48.72 &25.51 &63.60 \\
0.5 &67.77 &53.75 &{74.65} &60.63  &{93.55} &{84.65} &49.06 &25.48 &63.70 \\
0.7 &67.87 &53.69 &74.74 &60.56 &\textbf{93.57} &\textbf{84.70}  &49.26 &25.47 &\textbf{63.73} \\
0.9 &\textbf{67.94} &53.65 &\textbf{74.80}  &60.42 &93.54 &84.62 &\textbf{49.50}  &25.47 &\textbf{63.73} \\
\bottomrule
\end{tabular}
}
\vspace{-0.1in}
\caption{\small{}\textbf{Analysis of the attention scalar $\alpha$.} Results of RDC-FT with the attention scalar $\alpha=\{0,0.1,0.3,0.5,0.7,0.9\}$. $\alpha=0$ represents the RDC-FT results without attention strategy.}
\label{tab:alpha}
\vspace{-0.1in}
\end{table}

\subsection{Effect of the attention scalar $\alpha$}
The attention scalar $\alpha$  is used to increase the weights of the calibrated distance occurred in $\hat{R}$, here we investigate the effectiveness of different $\alpha$. 

The results in Tab.~\ref{tab:alpha} and Fig.~\ref{fig:parameters}(c)  show that this attention strategy can benefit the representation adaptation for FSL task in the target domain. In specific, moderately increasing the attention scalar ($\alpha$ from 0.1 to 0.5) can improve the effectiveness of the attention strategy. To the contrary, overly increasing the attention scalar ($\alpha$ from 0.5 to 0.9)  will introduce less even negative effect, resulting the decrease(slight increase) of the performance on 1(5)-shot learning. Therefore, the choice of $\alpha=0.5$ in the main paper is a moderate and robust parameter for the attention strategy.

\section{Visualisation}
\label{sec:visual}
\begin{figure}[t]
  \centering
   \includegraphics[width=0.95\linewidth]{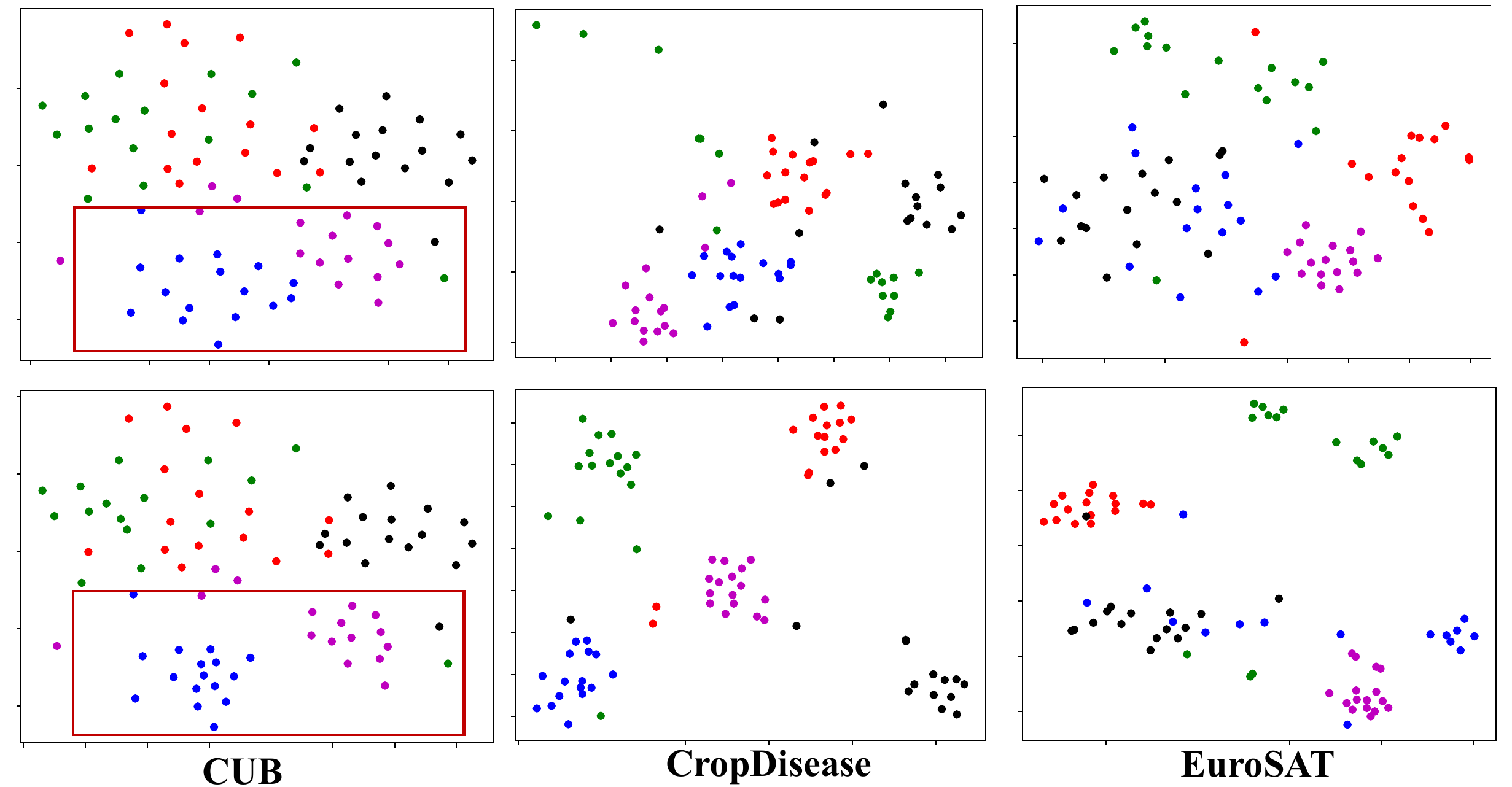}
     \vspace{-0.05in}
   \caption{\small{}\textbf{T-SNE visualisation of 5-way 1-shot tasks.} Different colours refer to different classes. We visualise the task features before (the 1st row) and after (the 2nd row) the RDC-FT method.}
  \vspace{-0.15in}
   \label{fig:visualisation}
\end{figure}

\camera{
To qualitatively show the effectiveness of our RDC and RDC-FT methods. We first show a case study of a FSL task from \textit{CUB} by comparing the original ranking list and the ranking list with RDC. As in Fig.~\ref{fig:visual_rdc}, for a given query data, our RDC method pulls the ground-truth support data closer to the query data, arriving at a more accurate position. This process is achieved by the calibration process of our RDC method.
For the RDC-FT method, we use t-SNE~\cite{van2008visualizing} to visualise the feature embeddings of FSL tasks randomly selected from target domains, \ie \textit{CUB}, \textit{CropDisease} and \textit{EuroSAT}. As in Fig.~\ref{fig:visualisation}, the feature representations with RDC-FT (in the 2nd row plots) have less within-class variations and large class margins compared to these without RDC-FT process (in the 1st row plots), showing that the RDC-FT method can guild a task-specific embedding where the samples can easily be classified by a simple NPC classifier.
Moreover, our RDC-FT method, as expected, is functioning as an implicit clustering process for FSL task. This can be qualitatively verified by the observation of the clustering effect as in the 2nd row plots of Fig.~\ref{fig:visualisation}.  
}

\begin{figure}[t]
  \centering
   \includegraphics[width=0.95\linewidth]{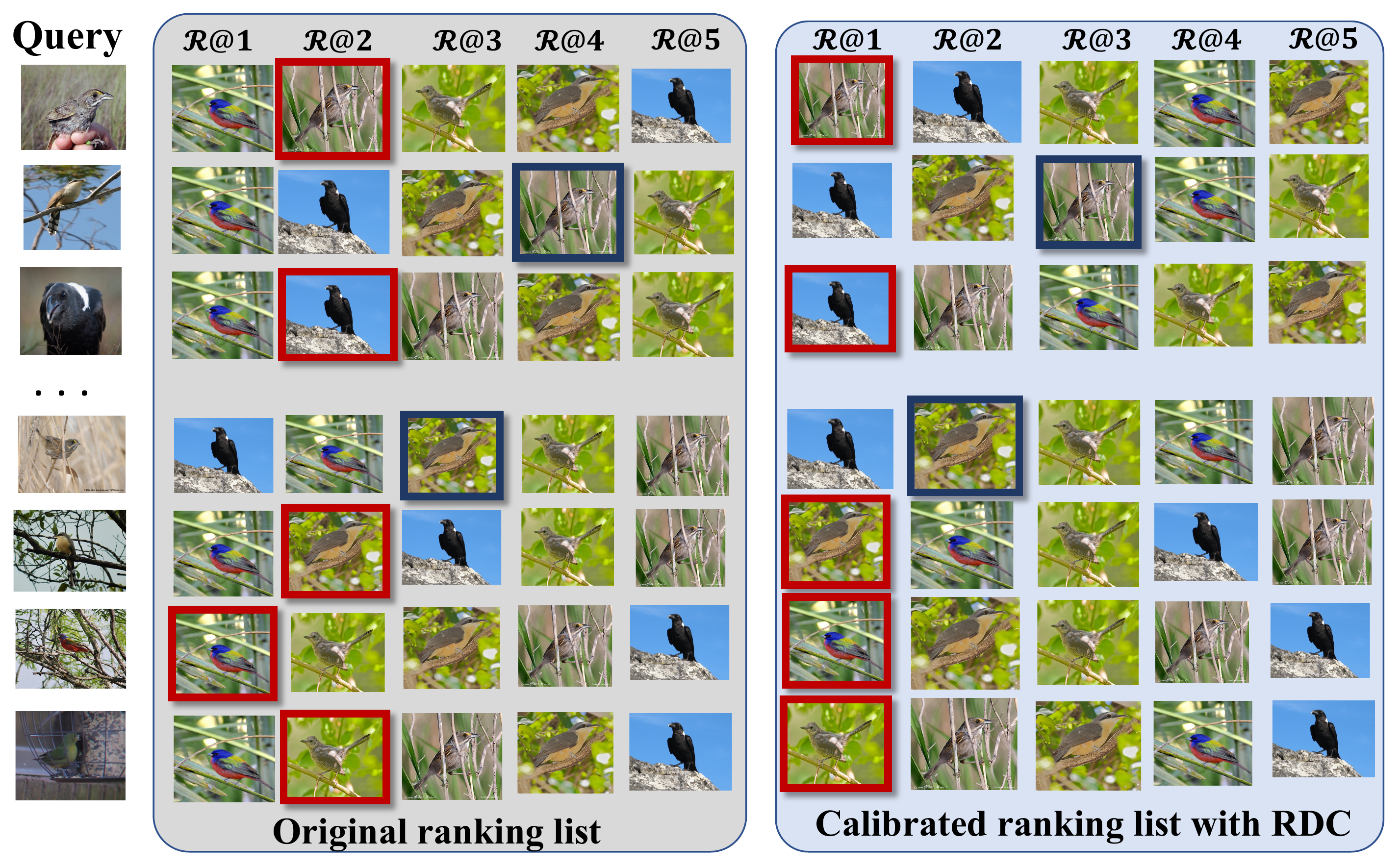}
   \vspace{-0.05in}
   \caption{\small{}\textbf{Ranking lists of a 5-way 1-shot task from \textit{CUB}.} The images with red/blue rectangle are the ground-truth support data for a given query. The RDC method calibrates the original ranking list to yield correct recognition results (the images with red rectangle) or closer pairwise distances (the images with blue rectangle).
       \label{fig:visual_rdc} }
  \vspace{-0.2in}
\end{figure}

\section{Details of Jaccard distance}
\label{sec:algorithm}
The Jaccard distance computing is an important part of RDC. In specific, the concept of Jaccard distance derives from~\cite{bai2016sparse} and the re-weighting strategy for Jaccard distance is also used in~\cite{zhong2017rerank}. We briefly introduce the computing process of Jaccard distance in the main paper. Here, we further illustrate more details for clearer description as in Algorithm~\ref{alg:algorithm}. 
In this pseudo-code, we illustrate the computing process of $k$-reciprocal discovery and encoding in line 3-9, and the discovery process and encoding process are presented in line 5-8 and line 9, respectively.
Then, the query expansion and Jaccard distance computing process are illustrated in line 11-14 of Algorithm~\ref{alg:algorithm}. 

\begin{algorithm*}[t]
\caption{{Jaccard distance computing}}
\label{alg:algorithm}
\KwData{pre-trained feature extractor $f_{\Phi}$; FSL task $\mathcal{T}$; $k$-nearest neighbors set: $k_1$, $k_2$.}
\KwResult{Jaccard distance $D_J=\{d_J(i,g_q),\; \mathrm{where} \; i,g_q \in [1,n]\}$.}
    Extract the embeddings for $\mathcal{T}$: $X=f_{{\Phi}}(\mathcal{T}), X\in \mathfrak{R}^{n\times m}$\;
    Compute original distances $D_o$ \;
    \tcc{ $k$-reciprocal discovery and encoding}
    \For{$i$ in $n$}
    {
    Compute the $k$-nearest neighbors ranking list $R_i(k)$ for $x_i$ \;
    \tcc{ $k$-reciprocal neighbor discovery process} 
    \For{$x_g$ in $R_i(k)$}
    {
    Compute the $k$-reciprocal nearest neighbors set $\mathcal{R}_g(k)$ for $x_g$  \;
    Expand $\mathcal{R}_g(k)$ by mining hard-positive samples: $\hat{ {R}}_i(k) \leftarrow {R}_i(k) \cup \mathcal{R}_g(\frac{1}{2}k) \; \;
 s.t. \left | {R}_i(k) \cap  {R}_g(\frac{1}{2}k)   \right |  \ge \frac{2}{3} \left |  {R}_g(\frac{1}{2}k) \right|$ \;
 }
    \tcc{ $k$-reciprocal encoding process} 
    Encode the expanded set $\hat{R}_i$ as $\mathcal{V}_i=[\mathcal{V}_{i,g_1}, \mathcal{V}_{i,g_2},...,\mathcal{V}_{i,g_n}]$ by  
        $\mathcal{V}_{i,g_q}=\begin{cases}
      e^{-d_o(x_i,x_{g_q})}& \text{ if } g_q\in\hat{\mathcal{R}}_i(k) \\
      0 & \text{otherwise.} 
    \end{cases}$ \;
    }
\tcc{ query expansion and Jaccard distance computing }
 \For{$i$ in $n$}
 {
    Expand the feature of $x_i$ as $\mathcal{V}_{i}=\frac{1}{|\hat{R}_i(k_2)|}{\textstyle \sum_{g_q\in \hat{R}_i(k_2)}\mathcal{V}_{g_q}}$ ; \ \tcp{ query expansion} 
   Compute the Jaccard distance $d_J(i,g_q)=1-\frac{|\hat{R}_i(k) \cap \hat{R}_{g_q}(k) |}{| \hat{R}_i(k) \cup \hat{R}_{g_q}(k)|}=
   1-\frac{ {\textstyle \sum_{j=1}^{n}}\mathrm{min}(\mathcal{V}_{i,g_j}, \mathcal{V}_{g_q,g_j}) }{{\textstyle \sum_{j=1}^{n}}\mathrm{max}(\mathcal{V}_{i,g_j}, \mathcal{V}_{g_q,g_j})}$ \;
   }
\end{algorithm*}

\begin{table}[t]
\centering
\scalebox{0.82}{
\begin{tabular}{ll}
\toprule
Symbol & Meaning \\
\midrule
$\mathcal{T}$       &FSL task in the target domain    \\
$x_i$       &Feature of $i$th sample in $\mathcal{T}$   \\
$D_o$        &Euclidean distance matrix in the original space     \\
$D_J$        &Jaccard distance matrix     \\
$\hat{D}_o$        &Calibrated distance matrix in the original space   \\
$\hat{D}_{sub}$        &Calibrated distance matrix in the subspace     \\
$\hat{D}_{com}$        &Complementary calibrated distance matrix     \\
$d_o(i,j)$   &Pairwise distance between $x_i$ and $x_j$          \\
$\mathbf{d}_o(i,:)$       &Pairwise distances between $x_i$ and $x_j \in \mathcal{T}$     \\
$d_J(i,g_q)$       &Jaccard distance between $x_i$ and $x_{g_q}$       \\
$R_i(k)$             &$k$-nearest neighbors ranking list of $x_i$        \\
$\hat{R}_i(k)$       &Expanded $k$-nearest neighbors ranking list of $x_i$         \\
$\mathcal{V}_{i,g_q}$  &Gaussian kernel of pairwise distance between $x_i$ and $x_{g_q}$       \\
\bottomrule
\end{tabular}
}
\vspace{-0.05in}
\caption{\small{Explanation of the symbols.}}
\label{tab:symbol}
\vspace{-0.05in}
\end{table}

\begin{table}[t]
\centering
\scalebox{0.85}{
\begin{tabular}{ll}
\toprule
\small{}Hyper-parameter & Meaning \\
\midrule
$k$        &Number of candidates in $R_i(k)$     \\
$k_2$      &Number of samples for updating $\mathcal{V}_i$     \\
$\lambda$  &Trade-off scalar to balance $D_o$ and $\hat{D}_{com}$    \\
$p$        &Dimensions of feature in the subspace      \\
$T$        &Number of epochs in fine-tuning stage     \\
$\tau$     &Temperature-scaling hyper-parameter         \\
$\alpha$   &Attention scalar \\
\bottomrule
\end{tabular}
}
\vspace{-0.05in}
\caption{\small{Explanation of the hyper-parameters.}}
\vspace{-0.1in}
\label{tab:hyper-parameters}
\end{table}

\section{Symbols and hyper-parameters}
\label{sec:symbol}
To clearly and fast understand the equations in the main paper, we list the symbols and hyper-parameters in the Tab.~\ref{tab:symbol} and Tab.~\ref{tab:hyper-parameters}, respectively.

\end{appendices}

\end{document}